\documentclass[11pt, letterpaper, logo, onecolumn, copyright]{gair}

\usepackage[authoryear, sort&compress, round]{natbib}

\usepackage[inkscapeformat=png]{svg}

\usepackage[most, breakable, skins]{tcolorbox}

\tcbuselibrary{skins}
\usepackage{lipsum}
\usepackage{tabularx}
\usepackage{afterpage}
\usepackage{booktabs}
\usepackage{subcaption}
\usepackage{makecell}
\usepackage{multirow}
\usepackage{multicol}
\usepackage{array}
\usepackage{float}
\usepackage{listings, listings-rust}
\usepackage{fontawesome5}
\usepackage{amssymb,graphicx}
\usepackage[dvipsnames]{xcolor}
\usepackage{hyperref}
\usepackage{cleveref}
\usepackage{longtable}
\usepackage{graphicx}
\usepackage{pdflscape}
\usepackage{adjustbox}
\usepackage{nicematrix}
\usepackage{CJKutf8}
\usepackage{ragged2e}

\usepackage{listings}
\usepackage{wrapfig}
\usepackage{xspace}
\usepackage{tikz}
\usepackage[normalem]{ulem}
\usepackage{hyperref}

\usepackage{pgfplots}
\usepackage{pgfplotstable}
\pgfplotsset{compat=1.18}

\definecolor{medgray55}{gray}{0.55}
\definecolor{medgray}{gray}{0.7}
\definecolor{litegray}{gray}{0.9}
\definecolor{gblue}{RGB}{210, 227, 252}
\definecolor{gred}{RGB}{250, 210, 207}
\definecolor{gyellow}{RGB}{254, 239, 195}
\definecolor{ggreen}{RGB}{206, 234, 214}
\definecolor{gorange}{RGB}{254, 223, 200}

\definecolor{gblue9}{RGB}{23, 78, 166}
\definecolor{gred9}{RGB}{165, 14, 14}
\definecolor{gyellow9}{RGB}{227, 116, 0}
\definecolor{ggreen9}{RGB}{13, 101, 45}
\definecolor{gorange9}{RGB}{176, 96, 0}

\definecolor{myblue}{rgb}{0,0,1}
\definecolor{myred}{rgb}{1,0,0}
\definecolor{mylightgray}{gray}{0.95}
\definecolor{myCite}{HTML}{643EAD}

\definecolor{highlightblue}{HTML}{185ABC}
\definecolor{cellHighlight}{HTML}{dbefff}

\newcommand{\textHL}[1]{{\color{gblue9}\detokenize{#1}}}
\newcommand{\mmwebpromax}{\texttt{MegaMath-Web-Pro-Max}}

\usepackage{minitoc} %

\noptcrule

\makeatletter

\renewcommand\paragraph{\@startsection{paragraph}{4}{\z@}%
            {-2.5ex\@plus -1ex \@minus -.25ex}%
            {1.25ex \@plus .25ex}%
            {\itshape\normalsize\bfseries}}
\makeatother
\newcolumntype{L}[1]{>{\raggedright\let\newline\\\arraybackslash\hspace{0pt}}m{#1}}
\newcolumntype{C}[1]{>{\centering}m{#1}}

\newcolumntype{R}[1]{>{\raggedleft\let\newline\\\arraybackslash\hspace{0pt}}m{#1}}

\definecolor{ao}{rgb}{0.0, 0.0, 1.0}

\newcommand\vcent[1]{\vcenter{\hbox{#1}}}

\newcommand\loudspeaker[1][3]{\ensuremath{\vcent{\rule{.6ex}{.6ex}}\kern-.5ex%
  \vcent{\scalebox{.6}[1]{\rotatebox[origin=center]{90}{$\blacktriangle$}}}%
  \ifnum#1>0\relax\kern.05ex\vcent{\scalebox{.4}{\ttfamily)}}%
  \ifnum#1>1\relax\kern-.4ex\vcent{\scalebox{.56}{\ttfamily)}}%
  \ifnum#1>2\relax\kern-.55ex\vcent{\scalebox{.7}{\ttfamily)}}%
  \fi\fi\fi}%
}

\newcommand{\para}[1]{%
  \par\noindent\textbf{#1}
}

\newcommand{\citeg}[1]{\cite[][\emph{inter alia}]{#1}}
\newcommand{\github}{\raisebox{-1.5pt}{\includegraphics[height=1.05em]{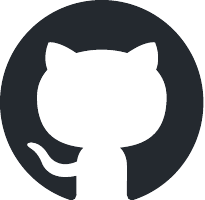}}\xspace}
\newcommand{\huggingface}{\raisebox{-1.5pt}{\includegraphics[height=1.05em]{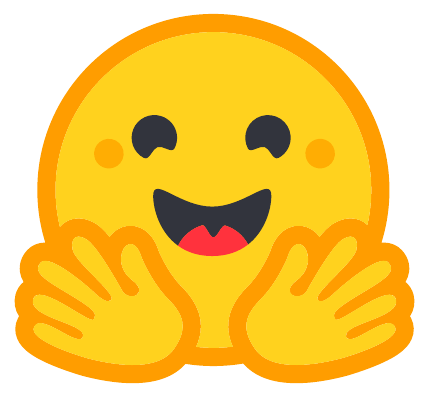}}\xspace}

\makeatletter
\renewcommand\subparagraph{%
 \@startsection {subparagraph}{5}{\z@ }{3.25ex \@plus 1ex
 \@minus .2ex}{-1em}{\normalfont \normalsize \bfseries }}%
\makeatother

\newcommand{\octo}{OctoThinker}

\let\cite\citep
\hypersetup{
  citecolor = myCite,  
  linkcolor = myCite,   
  urlcolor  = myCite
}

\title{\octo: Mid-training Incentivizes Reinforcement Learning Scaling}

\author{
    Zengzhi Wang\textsuperscript{*},
    Fan Zhou\textsuperscript{*},
    Xuefeng Li\textsuperscript{*},
    Pengfei Liu\textsuperscript{\ddag} \\
    Shanghai Jiao Tong University, SII, GAIR Lab \\ %
    \texttt{\{zengzhi.wang, zhoufan98, xuefengli, pengfei\}@sjtu.edu.cn} \\
    \vspace{1.5mm}
    \github \href{https://github.com/GAIR-NLP/OctoThinker}{\textbf{GAIR-NLP/OctoThinker}} ~ ~ ~ \huggingface \href{https://huggingface.co/OctoThinker}{\textbf{OctoThinker}} \\
    \vspace{-5mm}
}
\begin{abstract}
Different base language model families—such as Llama and Qwen—exhibit divergent behaviors during post-training with reinforcement learning (RL), especially on reasoning-intensive tasks. What makes a base language model suitable for reinforcement learning? Gaining deeper insight into this question is essential for developing RL-scalable foundation models of the next generation. In this work, we investigate how mid-training strategies shape RL dynamics, focusing on two representative model families: Qwen and Llama. 
Our study reveals that (1) high-quality mathematical corpora, such as \textbf{\texttt{MegaMath-Web-Pro}}, significantly improve both base model and RL performance, while existing alternatives (e.g., \textbf{\texttt{FineMath-4plus}}) fail to do so; (2) further adding QA-style data, particularly long chain-of-thought (CoT) reasoning examples, enhances RL outcomes, and instruction data further unlocks this effect; (3) while long-CoT improves reasoning depth, it can also induce verbosity of model responses and unstability of RL training, underscoring the importance of data formatting; (4) scaling mid-training consistently leads to stronger downstream RL performance. 
Building on these insights, we introduce a two-stage mid-training strategy—\emph{\textHL{Stable-then-Decay}}—in which base models are first trained on 200B tokens with a constant learning rate, followed by 20B tokens across three CoT-focused branches with learning rate decay. This yields \textbf{\texttt{\textHL{OctoThinker}}}, a family of models demonstrating strong RL compatibility and closing the performance gap with more RL-friendly model families, i.e., Qwen. We hope our work will help shape pre-training strategies for foundation models in the RL era. To support further research, we release our open-source models along with a curated math reasoning-intensive corpus of over 70 billion tokens (i.e., \mmwebpromax).

\end{abstract}

\begin{document}

\doparttoc
\faketableofcontents

\begingroup
  \renewcommand\thefootnote{}
  \footnote{\textsuperscript{*}Equal contribution.
            \textsuperscript{\ddag}Corresponding author.}
  \addtocounter{footnote}{-1}
\endgroup

\maketitle

\begin{tikzpicture}[remember picture,overlay,shift={(current page.north west)}]
\node[anchor=north west,xshift=15.5cm,yshift=-3.5cm]{\scalebox{0.7}[0.7]{\includegraphics[width=5.5cm]{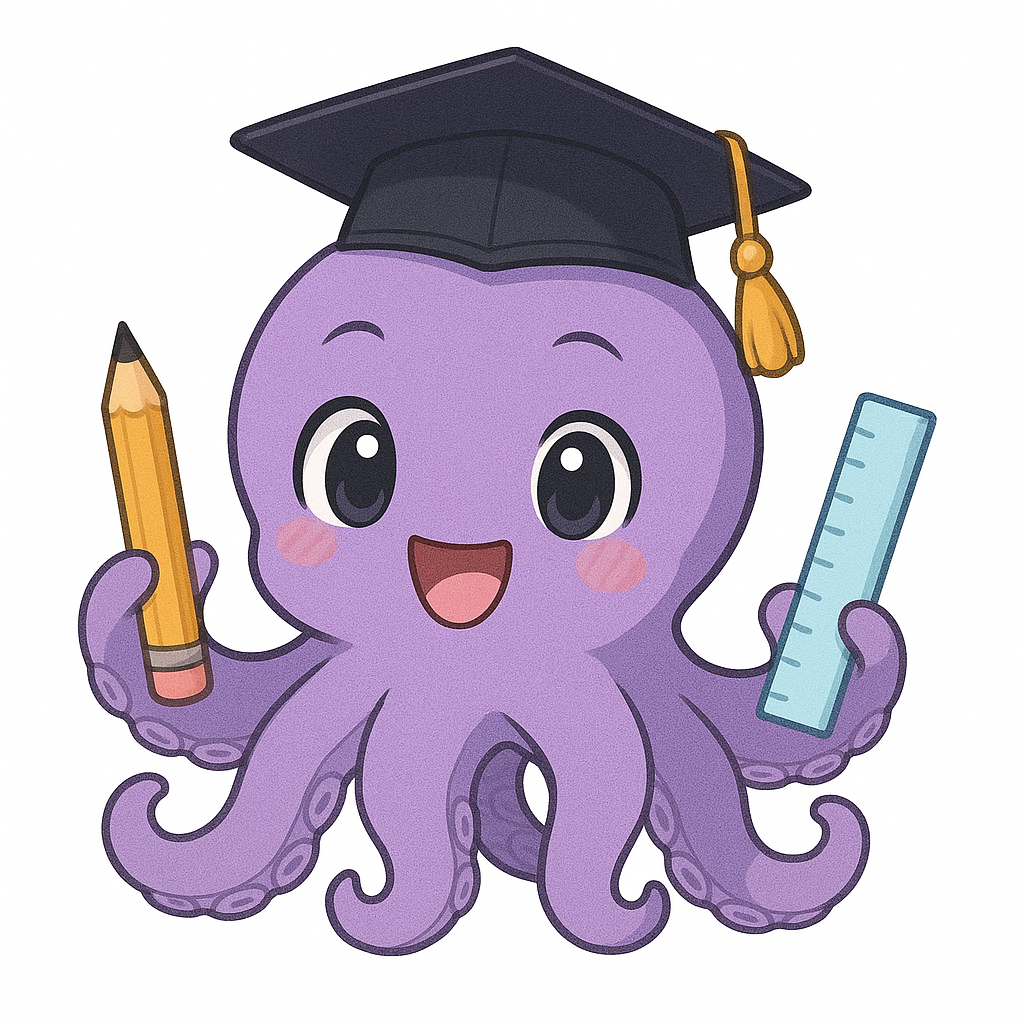}}};
\end{tikzpicture}

\begin{figure}[htbp]
\centering
\vspace{-30pt}
\includegraphics[width=0.81\linewidth]{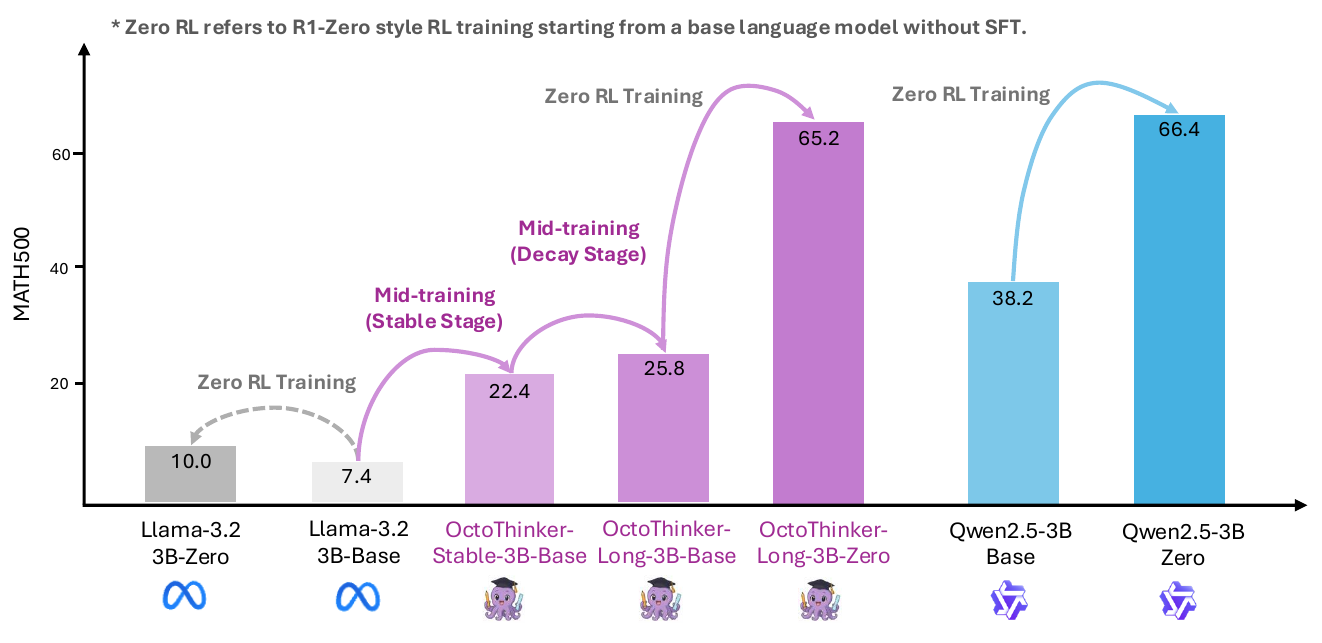}
\vspace{-4mm}
\caption{%
Our strategic mid-training incentivizes Llama's RL scaling, matching Qwen2.5 performance.
}
\label{fig:intro}
\vspace{-1mm}
\end{figure}

\section{Introduction}
\label{sec:intro}

Incentivizing large language models (LLMs) to think deeply through the chain of thought (CoT~\cite{DBLP:conf/nips/Wei0SBIXCLZ22-cot}) before giving the final answer with large-scale reinforcement learning (RL) is driving significant progress on the challenging reasoning tasks, i.e., solving competition-level mathematics problems, as demonstrated by OpenAI's o1~\cite{Contributors2024OpenAIo1} and o3~\cite{openai2025o3}. This also underscores a growing attention centered on RL as a means of boosting LLMs' reasoning performance. Deepseek-R1-Zero~\citep{guo2025deepseek} showcases a range of powerful and intriguing reasoning behaviors by directly applying large-scale RL to base language models, i.e., Deepseek-V3-Base~\citep{liu2024deepseek}. In line with this trend, several methods such as SimpleRL~\citep{zeng2025simplerl} and Open-Reasoner-Zero~\citep{hu2025open} have explored RL training on smaller base models—particularly the Qwen series~\citep{Yang2024Qwen25,yang2025qwen3}—achieving notable improvements in reasoning ability. However, despite these advances, replicating the success of R1-Zero-style training on other general-purpose base models, such as Llama series~\citep{grattafiori2024llama}, has proven difficult, also evidenced by recent studies~\citep{gandhi2025cognitive,liu2025understanding}. This naturally raises a fundamental question: \textbf{\emph{\textHL{What underlying factors cause the base models to exhibit divergent behaviors during RL training?}}} Understanding this could shed light on the scientific foundations that connect pre-training and the scalability of RL for reasoning, and may guide the design of future base models more amenable to reasoning-oriented RL.

In this work, we explore this question through the lens of mathematical reasoning and begin by observing a key difference in RL dynamics between two prominent model families: Qwen and Llama. Specifically, our preliminary studies reveal that Qwen models are much more amenable to RL scaling, while the Llama model tends to predict final answers prematurely and produce repetitive outputs during RL training. To better understand this discrepancy, we conduct a series of large-scale and controlled mid-training interventions on Llama models, followed by RL training. Our findings highlight that the quality of mathematical pre-training corpora is critical for successful RL performance. For instance, we find that \texttt{MegaMath-Web-Pro}~\cite{Zhou2025MegaMath} offers significantly greater benefits for RL scaling than corpora like \texttt{FineMath-4plus}~\cite{DBLP:journals/corr/abs-2502-02737-smollm2}. On top of a high-quality mathematical pre-training corpus, incorporating QA-style data yields further improvements, and introducing a small amount of instruction-following data helps enhance RL effectiveness even more. We also observe that injecting long CoT data during mid-training introduces instability into the RL phase. To address this, we refine the RL prompt and adopt a progressive maximum response length scheduler to stabilize training and ensure consistent behavior. To support large-scale mid-training, we also curate a reasoning-intensive mathematical corpus exceeding 70 billion tokens, namely \mmwebpromax, with data quality on par with \texttt{MegaMath-Web-Pro}. In extended mid-training experiments on this dataset—scaling up to 100 billion tokens—we observe that increasing the mid-training budget can lead to noticeable improvements in downstream RL performance. Interestingly, these gains are often not immediately reflected in the standard evaluations of the mid-trained base model, highlighting a gap between base model evaluation metrics and RL-stage capabilities.

\textbf{\emph{\textHL{Can we turn Llama into a foundation model well-suited for RL scaling by further scaling up its mid-training?}}} Building on the insights above, we explore this question by adopting a two-stage (\emph{\textHL{stable-then-decay}}) mid-training strategy. In the first stable stage, we train Llama models on a high-quality mixture of pre-training corpus for 200B tokens using a constant learning rate. In the second decay stage, we anneal the learning rate and introduce distinct data mixtures—short CoT, long CoT, and a hybrid of both—to mid-train three separate branches. These branches are later refined through RL training, equipping them with stronger reasoning capabilities. Inspired by the multi-armed nature of an octopus, we name this model family \texttt{\textHL{OctoThinker}}. Experiments across all model sizes and 14 mathematical reasoning benchmarks demonstrate the effectiveness of our approach: both stages of mid-training lead to substantial performance gains, especially the first stage, which consistently delivers 10–20\% improvement. Building on these stronger base models, subsequent RL training further boosts performance, with each branch showing distinctive behavior patterns. Notably, our models post-RL achieve performance on par with Qwen2.5 of the same size, effectively narrowing the gap between Llama and other RL-friendly model families. These results confirm the power of scaled-up, reasoning-intensive mid-training in transforming Llama into a suitable base model for RL scaling. To foster open research, we will release our curated pre-training data, models, and training scripts. As we enter the era of RL scaling, we are prompted to ask: \emph{What kind of foundation models do we need?} We believe this new phase brings unprecedented challenges for foundation models—and we hope \textHL{OctoThinker} offers a meaningful step toward the next generation of reasoning-capable AI systems.

\section{Preliminaries}
\label{sec:preliminaries}

We begin by identifying a key difference in RL dynamics between two prominent model families—Qwen and Llama—through the lens of mathematical reasoning. This observation offers a concrete and measurable foundation that grounds our systematic investigation.

\subsection{Experiment Setup}

\para{RL Setup} We perform our RL experiments based on the {\text{verl}}~\citep{sheng2024hybridflow} framework and utilize the \text{GRPO}~\citep{shao2024deepseekmath} algorithm.  For RL training prompts, we adopt the {\text{MATH8K}}~\footnote{\url{https://hf.co/datasets/hkust-nlp/SimpleRL-Zoo-Data/tree/main/simplelr_qwen_level3to5}} dataset due to its moderate difficulty and concise composition. We configure the global training batch size to 128, set the number of rollout responses per query to 16, and use a PPO mini-batch size of 64. The sampling temperature is set to 1.0, with a maximum output length of 4096 tokens. We use a learning rate of $1 \times 10^{-6}$ and set the KL loss coefficient to 0 in the verl configuration. Empirically, we find that setting the ratio between sampling and gradient updates to 2 leads to more stable RL training. Unless otherwise specified, we employ a simple prompt template of {\color{gblue9}``\verb|Question:{}\nAnswer:{}|''} to format training examples.

\para{Choices of Base Model} We employ Llama-3.2-3B-Base~\cite{DBLP:journals/corr/abs-2407-21783-llama-3} and Qwen2.5-3B-Base~\cite{Yang2024Qwen25} to perform R1-Zero styled RL training given the moderate model size.

\para{Evaluation} We adopt the few-shot prompting evaluation for base language models and employ zero-shot evaluation for RL-tuned models. Specifically, we employ GSM8K~\citep{cobbe2021gsm8k}, MATH500~\cite{DBLP:journals/corr/abs-2305-20050-verify-step-by-step}, OlympiadBench~\cite{DBLP:conf/acl/HeLBHTSHHHZLQL024-OlympiadBench}, and AMC23 as indicator tasks to analyze RL dynamics. To assess base model performance, we further include MATH~\citep{hendrycks2021measuring}, SAT-MATH~\citep{azerbayev2024llemma} , MathQA~\citep{amini2019mathqa}, MMLU-STEM~\cite{hendrycks2021measuring}, OCW Course~\cite{DBLP:conf/nips/LewkowyczADDMRS22-minerva}, MAWPS~\citep{koncel2016mawps}, SVAMP~\citep{patel2021nlp}, ASDiv~\citep{miao2021diverse}, and TabMWP~\citep{lu2023tabmwp}.

\subsection{Observations}

\begin{figure}[htbp]
\centering
\vspace{-12.5pt}
\includegraphics[width=0.85\linewidth]{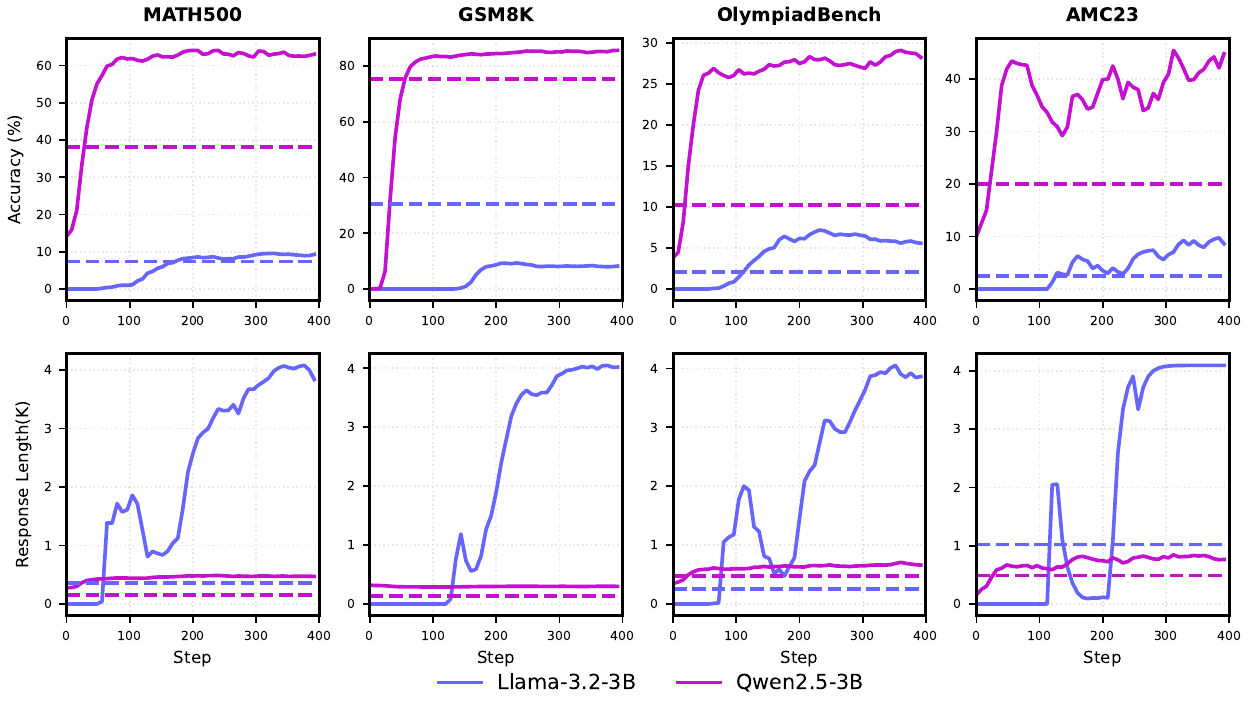}
\vspace{-1.5mm}
\caption{%
Training dynamics comparison (downstream performance and the average length of correct responses) between Llama-3.2-3B and Qwen2.5-3B. The dashed line indicates the few-shot evaluation performance and average length of correct responses of the corresponding base models.
}
\label{fig:analysis_motivation}
\end{figure}

During RL training on Llama-3.2-3B-Base and Qwen2.5-3B-Base, we observe notably different and intriguing training dynamics regardless of their performance, as shown in Figure~\ref{fig:analysis_motivation}. Specifically, the length of correct responses from the Qwen model increases steadily and reasonably throughout training, whereas Llama exhibits abnormal behavior—its average response length escalated dramatically, reaching up to 4,096 tokens. 

Upon closer inspection of the Llama model’s output, we find that it typically begins with {\color{gblue9}``\verb|\boxed:{}|''}, followed by extremely obvious repetition until hitting the max response length, in stark contrast to Qwen’s coherent and natural reasoning output. The evaluation results further highlight the divergence: The RL-tuned Qwen2.5-3B achieves substantial improvements over its base model across a wide spectrum of benchmarks, from simple to complex math reasoning tasks. 
Meanwhile, Llama-3.2-3B experiences only marginal gains—or even regressions, as seen on GSM8K—likely due to the distributional gap between the RL training set (e.g., MATH8K) and GSM8K. The above observations motivate us to attribute the reason to their potential divergence of pre-training despite their opaque details.

These observations also further prompt a more fundamental question: 
{\color{gblue9}\textit{Can we intervene in the Llama base language models via mid-training to make it more amenable to RL scaling?}}  Specifically, in this work, we aim to explore a range of mid-training intervention strategies—methods that adjust the pre-training trajectory of LLMs—to examine their downstream impact on RL dynamics.

\begin{tcolorbox}[
colback=gblue9!5, 
colframe=gblue9!75, 
coltitle=white,            
colbacktitle=gblue9!75,
title={\parbox{\textwidth}{\textbf{What is Mid-training?}}},
]
Mid-training is a mid-stage whose computational and data (token) requirements are intermediate between pre-training and post-training. It aims to achieve specific objectives—such as domain and language expansion~\citeg{Dou2025Sailor2}, long-context extension~\citeg{DBLP:journals/corr/abs-2404-14219-phi-3,DBLP:journals/corr/abs-2412-08905-phi-4}, improving data quality~\citeg{DBLP:journals/corr/abs-2404-06395-minicpm,OLMo2024ai2}, leveraging large-scale synthetic data~\citeg{Yang2024Qwen25,Yang2024Qwen25Math,yang2025qwen3}, and preparing for post-training, among others—by significantly altering data quality and distribution~\citeg{DBLP:journals/corr/abs-2407-21783-llama-3,DBLP:journals/corr/abs-2412-01253-yi-lightning} (and/or modifying model architecture to improve inference efficiency~
\citeg{DBLP:journals/corr/abs-2411-19146-puzzle,bercovich2025llama-nemotron}).\footnote{In the absence of a precise or widely agreed-upon definition, here, we aim to introduce a concise and rigorous definition of \emph{mid-training} within this context. The term was reportedly first mentioned in an OpenAI job description in mid-2024.  A detailed blog for this term is available at \url{https://vintagedata.org/blog/posts/what-is-mid-training}}
\end{tcolorbox}

\section{Digging Deeper: Exploring Key Factors through Controllable Mid-training}

\begin{figure}[htbp]
\centering
\includegraphics[width=0.88\linewidth]{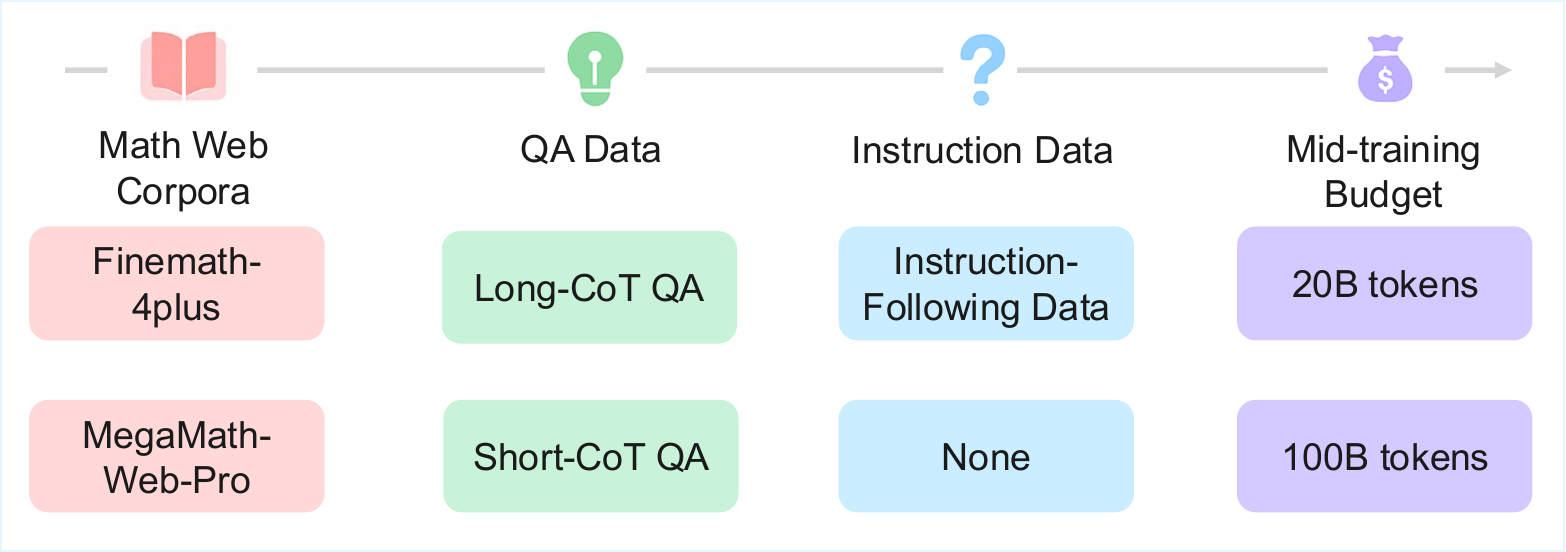}
\vspace{-1.5mm}
\caption{%
Potential factors in mid-training that could impact the post-training stage.
}
\label{fig:analysis_factors_config}
\vspace{-4.5mm}
\end{figure}

 We aim to investigate the impact of several factors during mid-training on RL performance through head-to-head experiments, as shown in Figure~\ref{fig:analysis_factors_config}. Specifically, we examine the effects of data quality of math web corpora, the inclusion or exclusion of QA-format data, the nature of the QA data itself, the presence of general instruction-following data in mid-training, as well as the pre-training token budget. These systematic analyses help us gain a deeper understanding of the connection between pre-training and RL dynamics and figure out suitable recipes for scaled-up mid-training.
 
\subsection{Experimental Setup}

\para{Mid-training setup} By default, we perform mid-training with Llama-3.2-3B-Base on diverse datasets and training configurations within a 20B-token training budget. We use a cosine learning rate scheduler without warmup, with a peak learning rate of 3e-5 and a minimum learning rate set to one-tenth of the peak. The default sequence length is 8,192, and the batch size is 4 million tokens. Training is conducted using the Nanotron framework.~\footnote{\url{https://github.com/huggingface/nanotron}}

\para{RL setup} We follow the exact same RL setup as described above in Section~\ref{sec:preliminaries}, unless stated otherwise.

\begin{table}[!ht]
\centering
\caption{%
    Statistics and Types of different datasets used in our experiments.~{{$^\diamond$}We use the \texttt{TULU3-sft-personna-instruction-following} subset.}
}
\label{tab:dataset_stats}
    \begin{NiceTabular}{L{7.0cm}|L{5.5cm}|c}
    \toprule
    \textbf{Dataset}                                 & \textbf{Type}                                          & \textbf{\# Tokens (B)} \\ \midrule
    FineMath-4plus~\cite{DBLP:journals/corr/abs-2502-02737-smollm2}                           & \multirow{3}{=}{Math Web Documents}               & 9.57          \\
    MegaMath-Web-Pro~\cite{Zhou2025MegaMath}                        &                                                &13.00           \\
    \mmwebpromax ~(Ours)                   &                                                & 73.80        \\ \midrule
    MegaMath-QA~\cite{Zhou2025MegaMath}                              & QA (Short-CoT)                                 & 5.94          \\
    OpenR1-Math-220K~\cite{openr1}                         & QA (Long-CoT)                                  & 1.05          \\ \midrule
    TULU3-sft\textsuperscript{$\diamond$}~\cite{DBLP:journals/corr/abs-2411-15124-tulu3} & \multirow{3}{=}{General Instruction Following} & 0.01          \\
    WildChat~\cite{zhao2024wildchat}                                 &                                                & 0.29          \\
    UltraChat-220K~\cite{DBLP:conf/emnlp/DingCXQHL0Z23-ultrachat}                           &                                                & 0.51          \\ \bottomrule
    \end{NiceTabular}
\end{table}

\para{Datasets} The datasets used to support our controllable experiments are summarized in Table~\ref{tab:dataset_stats}. For the OpenR1 dataset, we concatenate the question and the thinking process enclosed within \texttt{<think>} and \texttt{</think>} using a line break. For the  general instruction following datasets, we only retain high-quality conversations, such as those derived from GPT-4, and formated the conversations as
``{\color{gblue9}\verb|User:{}\nAssistant:{}|}''.

\para{The Curation of \mmwebpromax} We curate \mmwebpromax~to support large-scale ablation studies and mid-training. The corpus is constructed using an efficient classifier to recall documents from MegaMath-Web~\citep{Zhou2025MegaMath},  followed by refinement using a powerful instruction-following LLM. Specifically, we uniformly and randomly sample millions of documents from the MegaMath-Web corpus, stratified by publication year, and annotate them using Llama-3.1-70B-instruct. Each document is graded for its usefulness in studying mathematics on a scale from 0 to 5 using a grading prompt (see Figure~\ref{app-fig:finemath-prompt}). We heuristically extract scores from the model's critiques: documents scoring below 3 were labeled as negative examples, while those scoring 3 or above were considered positive. We observe that existing classifiers, such as \texttt{finemath-classifier}, are highly sensitive to the choices of text extractors during data curation. This motivates us to train our own classifier, selecting \texttt{fasttext} for its efficiency. Consistent with the findings of ~\citet{Zhou2025MegaMath}, we find preprocessing to be critical for recall performance. Our preprocessing pipeline includes lowercasing text, filtering excessively long words, and removing line breaks and extraneous non-alphanumeric characters. 
\begin{wrapfigure}[18]{r}{0.6\textwidth}
\centering
\vspace{-12.5pt}
\includegraphics[width=0.95\linewidth]{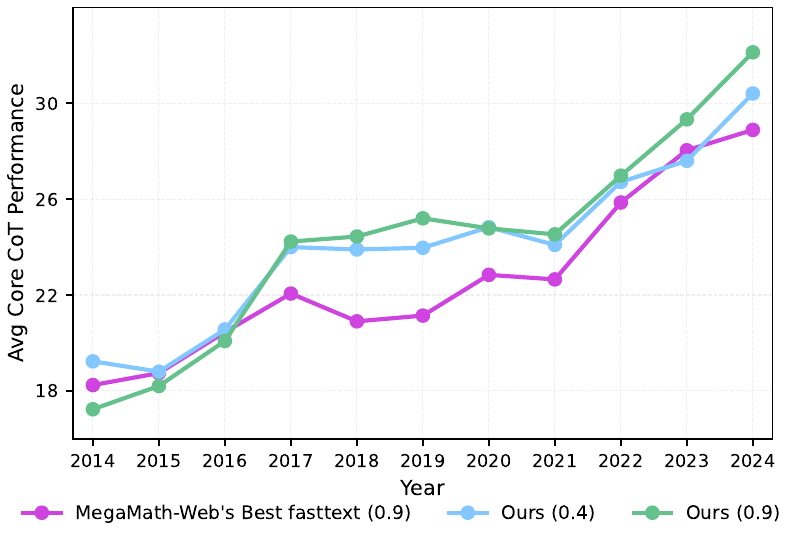}
\vspace{-1.5mm}
\caption{
Comparison between our fasttext-recalled corpus (w/o LLM refinement) and MegaMath-Web, following its yearly  dump comparison setup under a 5B-token pre-training budget.  Recall thresholds shown accordingly.
}
\label{fig:megamath_web_pro_max_data_quality_comparison}
\vspace{-4.5mm}
\end{wrapfigure}
Following MegaMath-Web’s yearly dump comparison setup, we evaluate the quality of our recalled corpus under different thresholds, as shown in Figure~\ref{fig:megamath_web_pro_max_data_quality_comparison}. The recall threshold controls the trade-off between data quantity and quality: a higher threshold (e.g., 0.9) yields better quality but retains fewer tokens. Finally, we select a threshold of 0.4. Given the noisy and poorly structured nature of many documents, we employ Llama-3.1-70B-instruct to refine the text using a prompt (see Figure~\ref{app-fig:text-refining-prompt}) inspired by MegaMath-Web-Pro. The resulting dataset, \mmwebpromax, contains approximately 5.5 times more tokens than MegaMath-Web-Pro. Empirical evaluations during pre-training indicate that \mmwebpromax~maintains comparable data quality, making it a strong candidate as a foundational corpus for large-scale mid-training. Besides, we also explored to supplement the positive seed set with (long)CoT examples from common math problem-solving datasets to improve the classifier's ability to recall reasoning-intensive content. However, this approach retained only around 20B tokens, which we deemed insufficient in scale and thus did not adopt.

\subsection{On the Inclusion and Data Quality of Math Web Corpora}

\begin{figure}[htbp]
\centering
\includegraphics[width=0.85\linewidth]{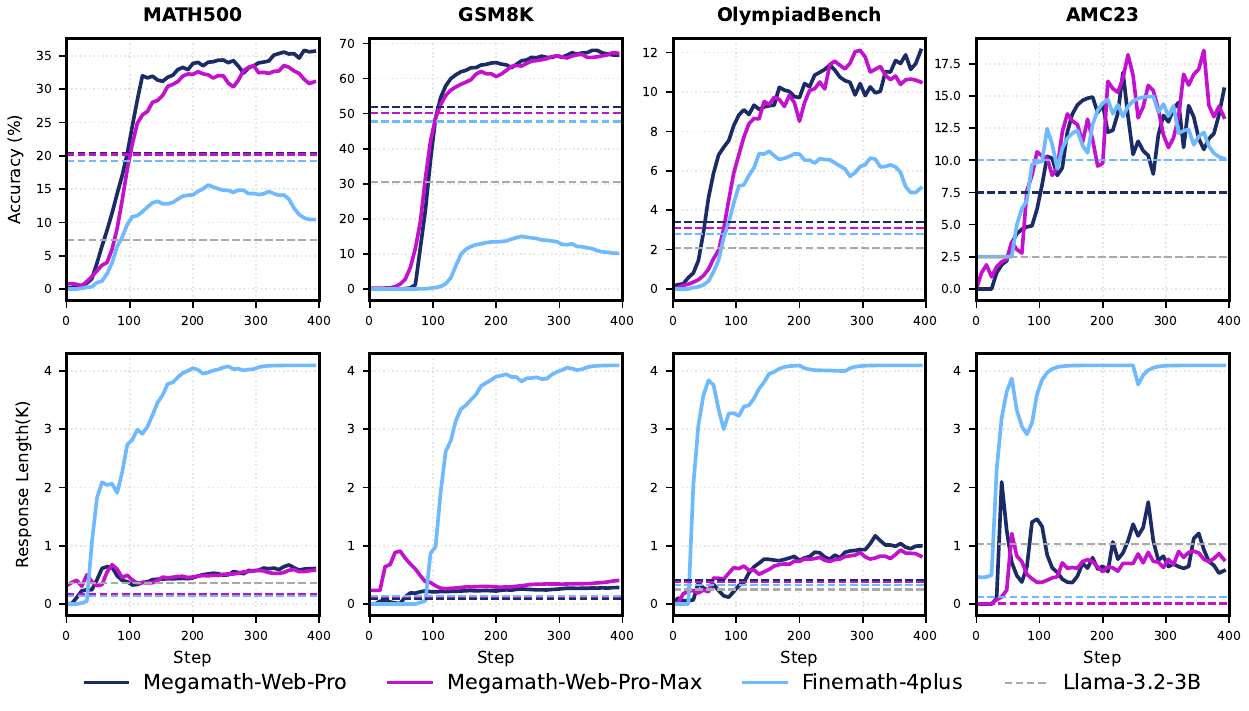}
\vspace{-1.5mm}
\caption{%
The effect of different math web corpora during mid-training. We performed mid-training on each corpus with a 20B-token training budget.
}
\label{fig:analysis_web_data_quality}
\end{figure}

Web corpora provide a solid foundation during pre-training. We believe that math-specific web corpora, along with their data quality, continue to play a crucial role during mid-training. We begin our systematic analysis by performing mid-training on different math web corpora and holding other factors being constant. As shown in the Figure~\ref{fig:analysis_web_data_quality}, mid-training on math web data improves performance over the base model, with MegaMath-Web-Pro and \mmwebpromax~showing slightly better gains than Finemath-4plus.
After RL training, we find that mid-training on math web corpora improves RL performance to varying degrees. MegaMath-Web-Pro and \mmwebpromax~bring significant gains for Llama in RL training, while Finemath-4plus yields only marginal improvements—highlighting the clear differences in data quality. Furthermore, we observe that models trained on FineMath-4plus exhibited abnormal behavior, with response lengths rapidly increasing until reaching the maximum limit of 4,096 tokens. The outputs typically begin with ``{\color{gblue9}\verb|\boxed{}|}'' and devolve into repetitive ``Solution'' statements.
Given these observations, we select MegaMath-Web-Pro as our default mathematical corpus and also \mmwebpromax~for scaled mid-training.

\begin{tcolorbox}[
colback=gred!25, 
colframe=gred!95, 
]
\textbf{Takeaway:} High-quality math pre-training corpora play a dominant role in RL scaling. We finally adopt MegaMath-Web-Pro and our curated \mmwebpromax~in this work. 
\end{tcolorbox}

\subsection{On the Inclusion and Nature of QA-Format Data}

\begin{figure}[htbp]
\centering
\vspace{-12.5pt}
\includegraphics[width=0.85\linewidth]{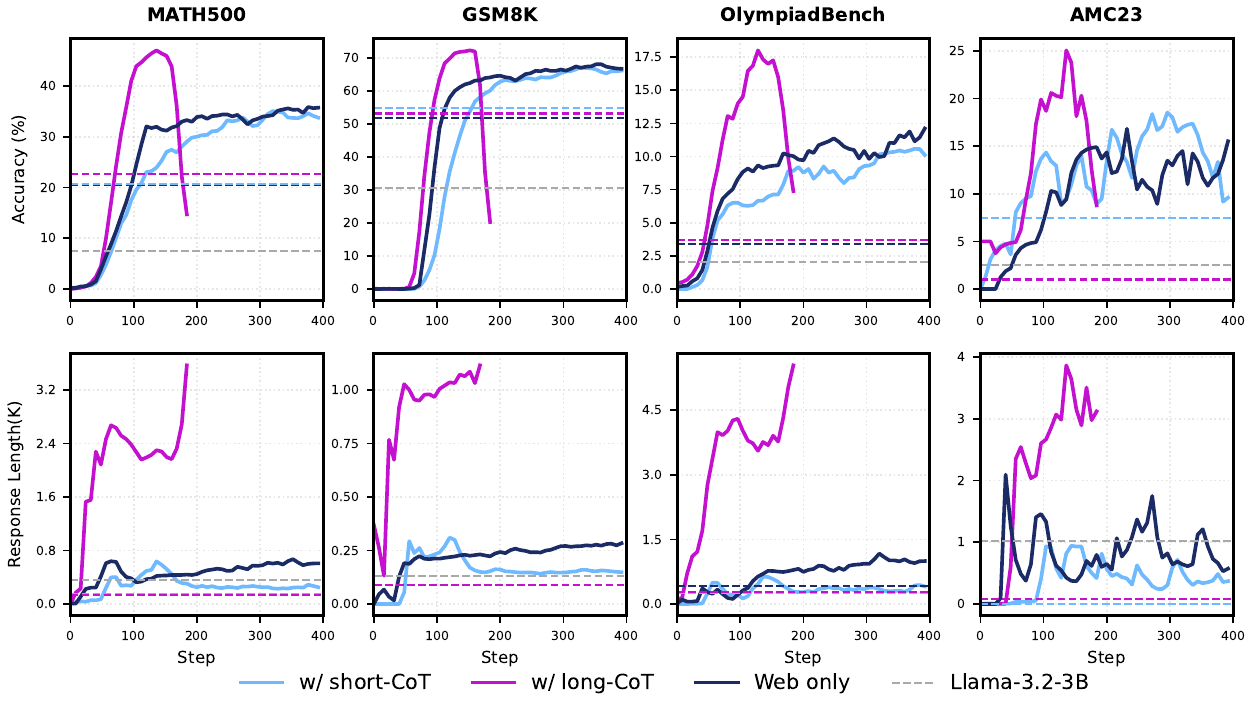}
\vspace{-1.5mm}
\caption{%
Impact of incorporating CoT data with varying characteristics during mid-training (9:1 mixture ratio). The figure also illustrates performance and average lengths of correct responses for Llama-3.2-3B-Base and its mid-trained variants for reference (in dashed line with different colors).
}
\label{fig:analysis_qa}
\vspace{-2.5mm}
\end{figure}

Intuitively, introducing QA data into pre-training and mid-training improves model performance, as previously examplified in \citet{bi2024deepseek} and \citet{hu2024minicpm}.
We further investigate this using a 9:1 web-to-QA data mix. 
We hypothesize that QA data's short Chain-of-Thought (short-CoT, from MegaMath-QA) and long-CoT (from OpenR1-Math-220K) reasoning, which may include self-reflection and backtracking, enhance base model performance and RL training. Maximum response lengths were 8,192 tokens for long-CoT models and 4,096 for others.

As shown in Figure~\ref{fig:analysis_qa}, incorporating QA data into mid-training generally yields performance gains for the base model, though these gains are marginal, as indicated by dashed lines. 
After RL training, incorporating short-CoT data into mid-training shows no improvements compared to mid-training on web data alone, possibly due to the data distribution gap (see \S~\ref{sec:octothinker-decay-stage} for more ablation studies), while long-CoT data brings significant performance gains. 
However, incorporating long-CoT data introduces challenges with unstable RL training, evidenced by sudden performance drops and sharp increases in response length. 
We also explore methods for stabilizing RL training, which we discuss in the following sections.
\begin{tcolorbox}[
colback=gred!25, 
colframe=gred!95,
]
\textbf{Takeaway:} QA data could aid RL scaling, but gains depend on its distribution gap with downstream tasks. Long CoT patterns often induce excessive responses and sudden performance drops in RL-tuned models.
\end{tcolorbox}

\subsection{On the Inclusion of Instruction-following Data}
Incorporating instruction-following data into earlier-stage training has become an increasingly common practice. 
Works such as MiniCPM~\citep{hu2024minicpm} demonstrate that including high-quality unlabeled data and instruction-following data significantly improves downstream performance. 
We believe this inclusion is critically important for enhancing the base model's ability to follow instructions, which may be a potential key determining factor for successful RL training. 
We incorporate instruction-following data alongside web data and QA data in a 1:89:10 ratio.  For this, we combine these high-quality datasets with appropriate filtering and formatting: TULU3-sft-personas-instruction-following~\citep{lambert2024t}, WildChat~\citep{zhao2024wildchat}, and UltraChat-200K~\citep{ding2023enhancing}, totaling approximately 0.8B tokens.

\begin{figure}[htbp]
\centering
\vspace{-4.5pt}
\includegraphics[width=0.85\linewidth]{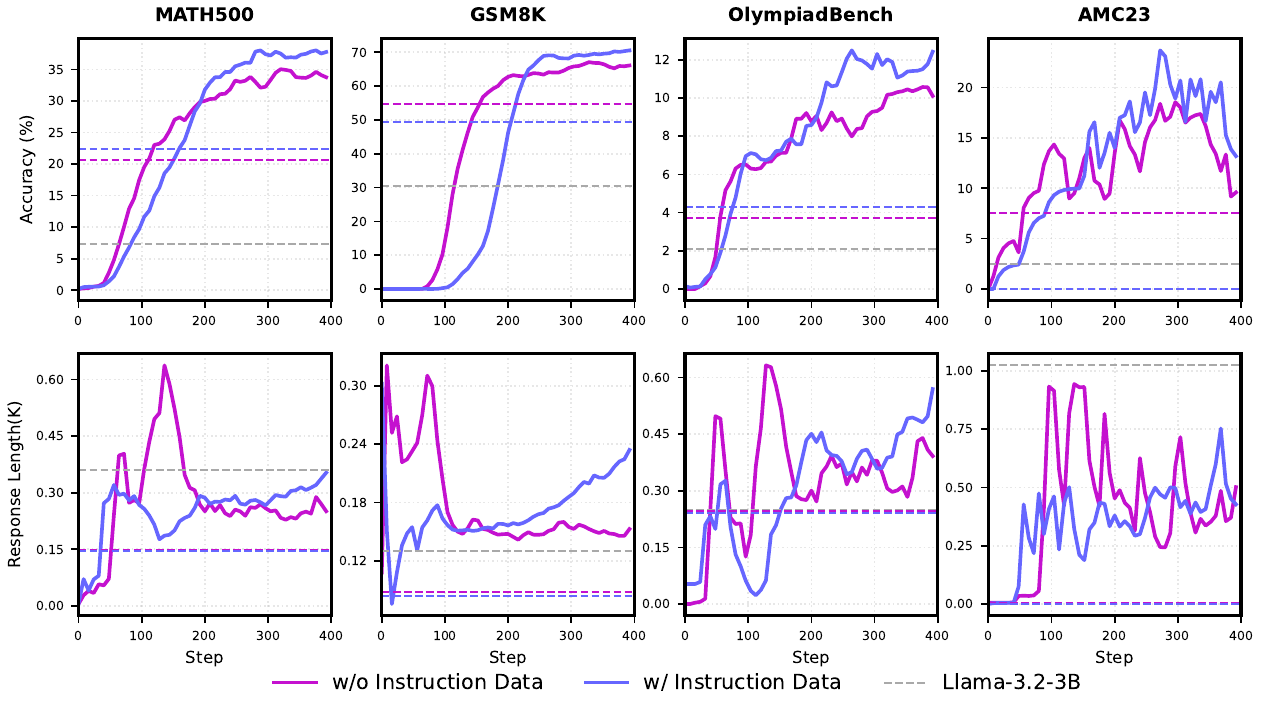}
\vspace{-1.5mm}
\caption{%
Impact of incorporating instruction-following data during mid-training with a mixture of web, short-CoT and instruction data in a ratio of 89: 10: 1 . The maximum response length is 4,096. The figure also illustrates performance and average lengths of correct responses for Llama-3.2-3B-Base and its mid-trained variants for reference (in dashed line with different colors).
}
\label{fig:analysis_instruction_into_short_mixture}
\vspace{-2.5mm}
\end{figure}

\para{Incorporating instruction-following data into the short-CoT mid-training mixture.}
As shown in Figure~\ref{fig:analysis_instruction_into_short_mixture}, after RL training, incorporating instruction-following data, unlocks the potential of short-CoT data, showing performance advantages over the exclusion case after 200 steps. 
Additionally, this inclusion helps stabilize response length, resulting in smoother increases compared to when instruction-following data is excluded.

\begin{figure}[!h]
\centering
\includegraphics[width=0.85\linewidth]{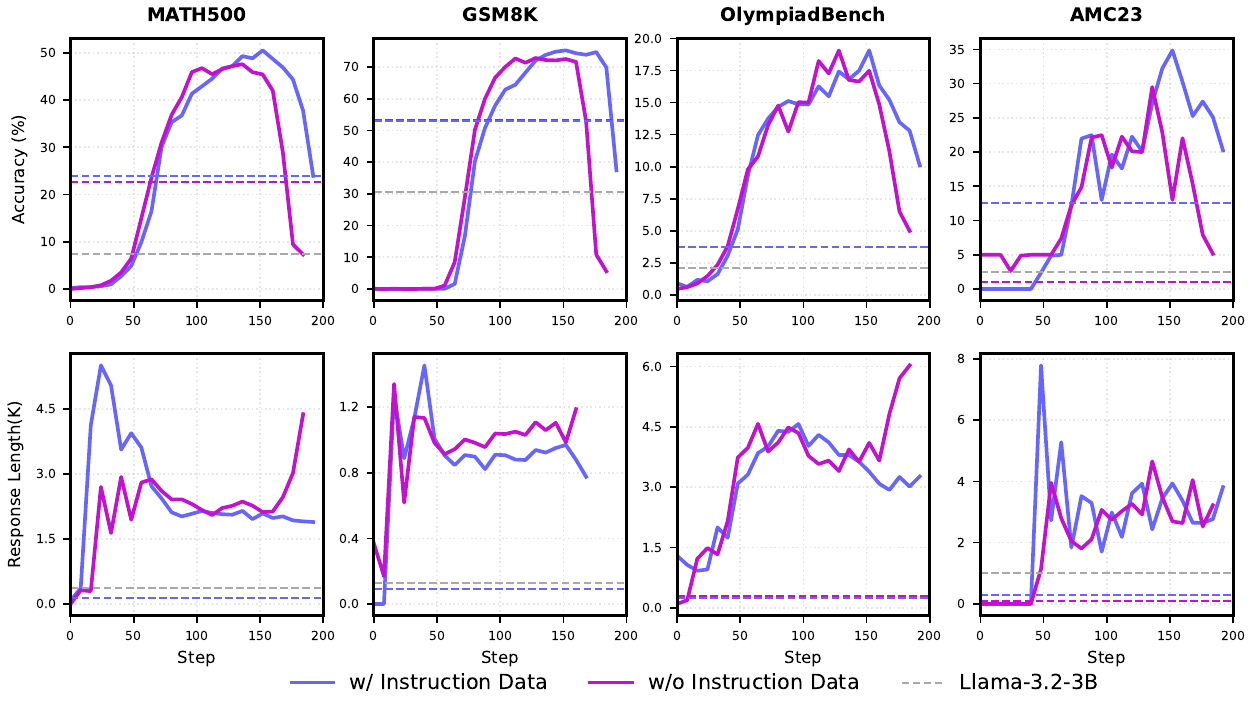}
\vspace{-1.5mm}
\caption{%
Impact of incorporating instruction-following data during mid-training with a mixture of web, long-CoT and instruction data in a ratio of 89: 10: 1. The maximum response length is 8,192. The figure also illustrates performance and average lengths of correct responses for Llama-3.2-3B-Base and its mid-trained variants for reference (in dashed line with different colors).
}
\label{fig:analysis_instruction_into_longcot_mixture}
\vspace{-4.5mm}
\end{figure}

\para{Incorporating instruction-following data into the long-CoT mid-training mixture.}
Similar to the challenges encountered earlier in RL training with the long-CoT mid-trained base model, as shown in Figure~\ref{fig:analysis_instruction_into_longcot_mixture}, incorporating instruction-following data shows performance improvements after 150 steps. However, this addition still fails to prevent the overall decline in RL performance and the rapid increase in response length. Note that we set the maximum response length to 8,192 tokens for these experiments.

Given the challenges encountered during RL training on the base model mid-trained on long-CoT data, we explore strategies to stabilize RL training by modifying the RL prompt template and maximum length scheduler.

\begin{figure}[!h]
\centering
\includegraphics[width=0.85\linewidth]{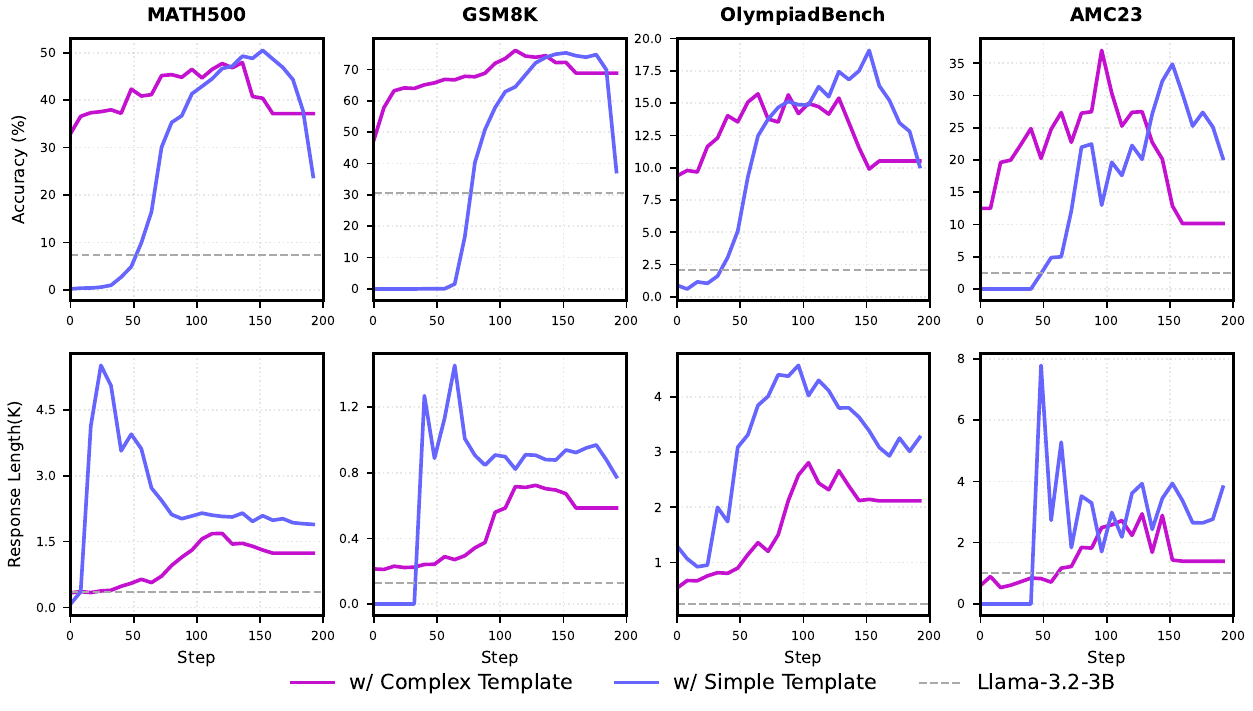}
\vspace{-1.5mm}
\caption{
Impact of different RL prompt templates. The figure also illustrates performance and average lengths of correct responses for Llama-3.2-3B-Base (in dashed line with different colors).
}
\label{fig:analysis_rl_prompt_templates}
\vspace{-2.5mm}
\end{figure}

\para{Effect of RL prompt template} The default template is ``{\color{gblue9}\verb|Question:{}\nAnswer:{}|}'', which we refer to as ``\emph{Simple Template}''. Here, we introduce an alternative, the ``\emph{Complex Template}'', adapted from the prompt design in Open-Reasoner-Zero~\citep{hu2025open}:

\begin{tcolorbox}[
colback=gblue!25, 
colframe=gblue!95, 
]
\textbf{Prompt Template: }A conversation between User and Assistant. The user asks a question, and the Assistant solves it. The assistant first thinks about the reasoning process in the mind and then provides the user with the answer. User: You must put your answer inside \verb|\\boxed{}| and your final answer will be extracted automatically by the \verb|\\boxed{}| tag.

\verb|{{prompt}}|

Assistant:
\end{tcolorbox}

We also control the maximum response length as 8,192 tokens. As shown in Figure~\ref{fig:analysis_rl_prompt_templates}, we find this  \emph{complex template} could clearly stabilize RL training compared to the \emph{simple template}, as evidenced by a smoother, more gradual increase in mean response length, as opposed to the sharp spikes observed with the simple template. Despite this stabilization, performance across evaluation benchmarks still deteriorates during the later stages of RL training, indicating need more exploring. Note that we adopt the \emph{complex template} as the default for all subsequent RL experiments.

\begin{figure}[!h]
\centering
\vspace{-6.5pt}
\includegraphics[width=0.85\linewidth]{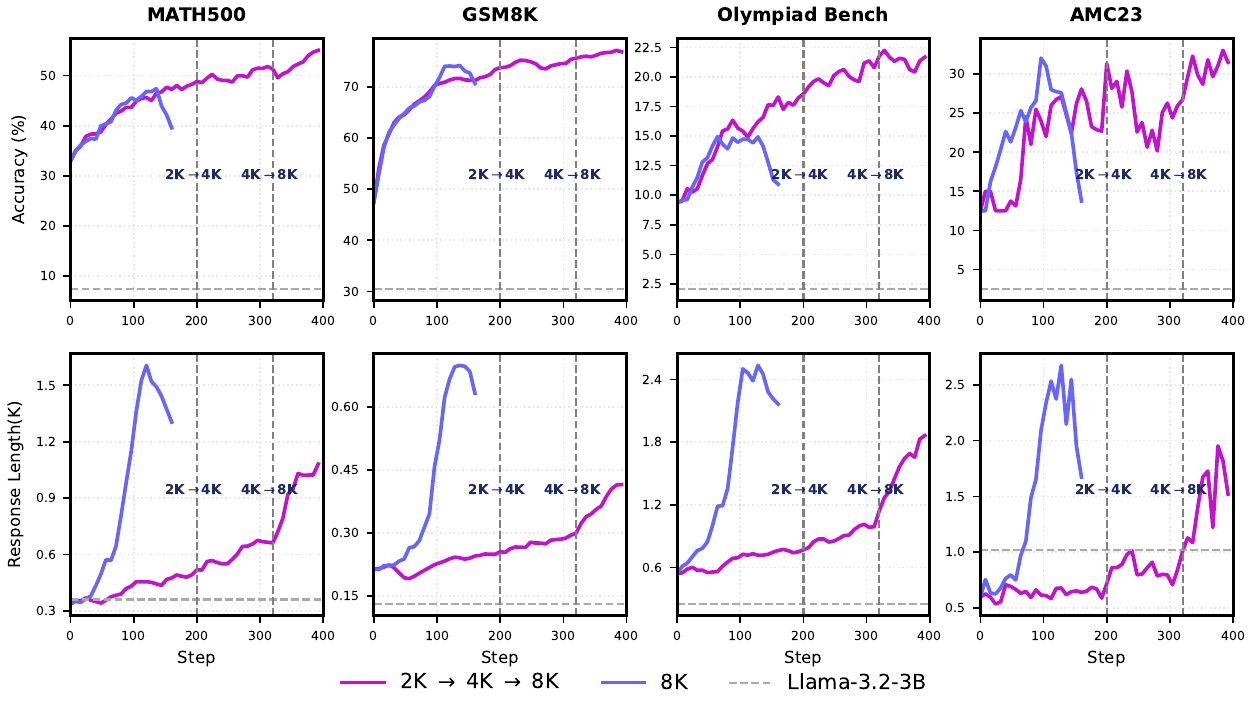}
\vspace{-1.5mm}
\caption{
Impact of the maximum length scheduler on the model response. The figure also illustrates performance and average lengths of correct responses for Llama-3.2-3B-Base in a dashed line.
}
\label{fig:analysis_max_response_length_scheduler}
\end{figure}

\para{Effect of the maximum response length} The default maximum context length is set to 8,192 tokens for long-CoT mid-trained models. Intuitively, we can delay the sharp rise in response length by gradually increasing the maximum response length in multiple stages. Specifically, we start with a limit of 2,048 tokens for the first 200 steps, increase it to 4,096 tokens from step 200 to step 320, and then further expand to the full 8,192-token context length from step 320 to step 400. As shown in Figure~\ref{fig:analysis_max_response_length_scheduler}, this progressive scheduling strategy significantly stabilizes RL training up to 400 steps, while consistently improving performance across benchmarks. In addition, the response lengths grow steadily and appropriately, highlighting the effectiveness of the progressive length scheduler.

\begin{figure}[!h]
\centering
\includegraphics[width=0.85\linewidth]{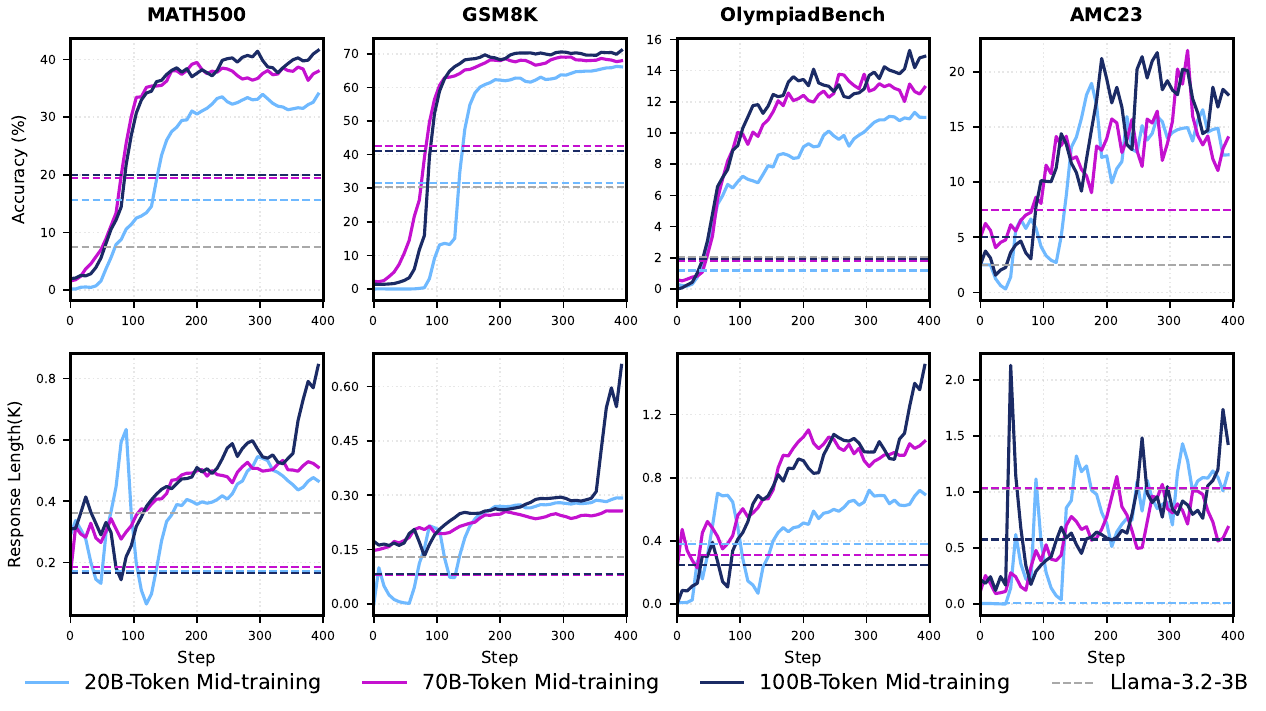}
\vspace{-1.5mm}
\caption{
Impact of scaling up the mid-training budget. The figure also illustrates performance and average lengths of correct responses for Llama-3.2-3B-Base and its mid-trained variants for reference (in dashed lines with different colors).
}
\label{fig:analysis_mid_training_token_budget}
\vspace{-4.5mm}
\end{figure}

\begin{tcolorbox}[
colback=gred!25, 
colframe=gred!95,
]
\textbf{Takeaway:} Introducing a small amount of instruction-following data can help unlock the potential of QA data and mitigate RL training collapse caused by long CoT. We address this issue by modifying the RL prompt template and applying a progressive maximum response length scheduler.
\end{tcolorbox}

\subsection{On the Issue of Mid-training Budget}

Could further scaling up mid-training improve RL performance? To explore this, we conduct a 100B-token mid-training run on \mmwebpromax using a default cosine learning rate scheduler. We select three intermediate checkpoints—trained on 20B, 70B, and 100B tokens, respectively—and perform RL training. When evaluating the base models, we observe that the 70B and 100B checkpoints achieved comparable performance, both significantly outperforming the 20B model. After RL training, interestingly, we find that increasing the mid-training token count consistently leads to improvements on RL performance despite varying degrees, whether moving from 20B to 70B or from 70B to 100B tokens. These findings highlight the importance of further scaling up the mid-training budget to unlock additional gains in downstream RL performance.

\begin{tcolorbox}[
colback=gred!25, 
colframe=gred!95,
]
\textbf{Takeaway:} Increasing the mid-training budget can improve RL performance, even if such gains are not evident in base model evaluations.
\end{tcolorbox}

\section{\octo-Base: Branching Reasoning Foundations via 2-Stage Mid-training}

Building upon the insights above, a natural question arises: \textbf{\emph{\textHL{Can we turn Llama into a foundation model well-suited for RL scaling by scaled-up mid-training?}}} 
We ultimately adopt a two-stage (\emph{\textHL{stable-then-decay}}) mid-training strategy to achieve both: 
(1) steady improvements in mathematical reasoning ability in the first stage;
(2) diversified model behaviors via branching in the second decay stage. 
Multi-stage pre-training has been validated as effective in prior work~\cite{hu2024minicpm,OLMo2024ai2}. The \emph{\textHL{stable-then-decay}} setup offers flexibility: the decay phase can begin at any point, enabling checkpoint selection independent of a fixed schedule. 
This also supports fair comparisons across different mid-training configurations. Importantly, decaying the learning rate in the second stage amplifies the effect of injected data, helping shape model behaviors more efficiently. Since the decay stage used for shifting model behaviors (in other words data distribution) is typically shorter, this approach also reduces the overall training cost in general. We name this resulting model family \texttt{\octo}\footnote{~~``\emph{\textHL{Octo}}'' is derived from ``octopus,'' symbolizing our base model family, which branches into variants trained with different strategies. ``\emph{\textHL{Thinker}}'' reflects the model’s final stage—reinforcement learning—where it is trained to think and reason, exhibiting frequent self-reflection and strong reasoning capabilities.}, inspired by the octopus’s multi-armed structure, reflecting its multiple branches.

\subsection{Recipe for the First Stage: Building Strong Reasoning Foundations}
Although the previous analysis has revealed several factors that are critical to building strong reasoning models, our mid-training resource table (see Table~\ref{tab:dataset_stats}) clearly shows that truly high-quality tokens are still scarce at this moment. Therefore, in the first phase, we adopt a relatively conservative strategy—primarily relying on high-quality web corpora such as \mmwebpromax and DCLM-Baselines~\citep{li2024datacomp}, supplemented with a small portion of synthetic data—to enable the model to improve steadily at scale. Following the training settings used in MegaMath-Llama~\citep{Zhou2025MegaMath}, we reduce the proportion of synthetic data and adopt a WSD-style~\citep{hu2024minicpm} learning rate scheduler, replacing the cosine learning rate with a constant learning rate and training for 200B tokens. We provide specific training configurations, i.e., data mixture and training hyper-parameters of the first-stage in Table~\ref{tab:octo-stable-data-mix} and Table~\ref{tab:octo-stable-train-config}. We refer to the resulting mid-training models as \texttt{OctoThinker-Base-Stable}.

\begin{table}[htbp]
\centering
\begin{minipage}[t]{0.48\textwidth}
    \centering
\caption{Dataset composition and weights in the first-stage.}
\label{tab:octo-stable-data-mix}
    \begin{NiceTabular}{l|r}
    \toprule
    \textbf{Dataset} & \textbf{Weight} \\
    \midrule
    DCLM-Baseline & 0.10 \\
    MegaMath-Web-Pro-Max & 0.725 \\
    MegaMath-Code & 0.0125 \\
    MegaMath-QA & 0.05 \\
    MegaMath Trans. Code & 0.0125 \\
    MegaMath Text Code Block & 0.10 \\
    \bottomrule
    \end{NiceTabular}
\end{minipage}
\hfill
\begin{minipage}[t]{0.48\textwidth}
    \centering
    \caption{hyper-parameters in stable stage.}
    \label{tab:octo-stable-train-config}
    \scalebox{0.9}{
    \begin{NiceTabular}{l|c}
        \toprule
        \textbf{Hyper-parameter} & \textbf{Llama-3.2-1B / 3B / 8B} \\
        \midrule
        \multirow{1}{*}{Context Length} & {8,192} \\
        \multirow{1}{*}{Batch Size} & {512} \\
        Max Steps & 50,000 \\
        \multirow{1}{*}{Warmup Steps} & {0} \\
        \multirow{1}{*}{Weight Decay} & {0.1} \\
        \multirow{1}{*}{Optimizer} & {AdamW} \\
        \multirow{1}{*}{LR Scheduler} & {Constant} \\
        \multirow{1}{*}{Learning Rate (LR)} & \makecell{5e-5/2e-5/1e-5} \\
        \bottomrule
    \end{NiceTabular}}
\end{minipage}
\end{table}

\subsection{Branching at the Second Stage: Seeking Perfect Blend for RL Scaling}
\label{sec:octothinker-decay-stage}

\subsubsection{Pilot Studies}

Building on prior experiments, we identify dataset quality and quantity as key drivers of effective mid-training and strong base model development. Before entering the decay stage, we conduct a series of controlled 10B-token mid-training experiments on the OctoThinker-3B-Base-Stable model—each followed by RL training—to investigate how different QA datasets affect downstream performance.

\para{Data Composition and Its Impact on RL}
We experiment with three QA datasets—MegaMath-QA, OpenR1-Math-220K, and OpenMathInstruct-2 (OMI2)—in varying proportions ($10$\%, $20$\%, $30$\%, and $40$\%) while holding constant $5$\% DCLM-Baselines data, $10$\% instruction data, and the remainder from MegaMath-Web-Pro. Ablation studies (see Figure~\ref{fig:decay_ablation_qa_ratio} in the Appendix) reveal that the origin of QA data plays a critical role. Specifically, OpenR1-Math-220K and OMI2 are derived from structured downstream datasets (e.g., GSM8K, MATH), while MegaMath-QA is sourced from less curated web documents. These differences in data source and distribution substantially impact downstream RL performance, highlighting the importance of distributional alignment between mid-training data and downstream tasks. In light of this, we adopt OpenMathInstruct-2, OpenR1-Math-220K (and further adopt the a-m-team’s distilled dataset~\footnote{\url{https://huggingface.co/datasets/a-m-team/AM-DeepSeek-Distilled-40M}}), and NuminaMath-1.5~\footnote{\url{https://huggingface.co/datasets/AI-MO/NuminaMath-1.5}} as our primary QA datasets for the decay stage, due to their closer resemblance to competition-style, reasoning-intensive benchmarks.

\para{Identifying the Optimal QA Ratio}
Across our ablation studies (also see Figure~\ref{fig:decay_ablation_qa_ratio}), we observe a consistent trend: increasing the QA data ratio leads to improved RL performance, which aligns with expectations due to the format similarity with RL objectives. However, gains begin to plateau beyond a 30\% QA mix, with 40\% showing diminishing returns across most benchmarks. We attribute this to token redundancy and lack of diversity at higher QA proportions. As a result, we adopt 30\% QA as the optimal ratio, balancing performance and data efficiency.

\begin{tcolorbox}[
colback=gred!25, 
colframe=gred!95,
]
\textbf{Takeaway:} 
The distribution gap between QA data and downstream tasks notably affects RL performance. Therefore, we introduce distribution-aligned QA data during the decay stage and find that a 30\% QA ratio offers the best trade-off between RL performance and QA data scale.
\end{tcolorbox}

\subsubsection{Final Decay Recipe}

For the decay stage, we explore two learning rate (LR) scheduler variants:
\begin{enumerate}
    \vspace{-3.5mm}
    \item \textbf{Constant LR decay}, where the LR remains fixed at 10\% of the final LR used in the stable stage.
    \item \textbf{Cosine decay to 10\%}, where the LR gradually decays to 10\% of the stable-stage final LR.
    \vspace{-3mm}
\end{enumerate}

\begin{wraptable}[10]{r}{0.475\textwidth}
    \centering
    \vspace{-5.5mm}
    \captionof{table}{Hyper-parameters for decay stage.}
    \label{tab:octo-decay-train-config}
    \resizebox{\linewidth}{!}{
        \begin{NiceTabular}{l|c}
            \toprule
            \textbf{Hyper-parameter} & \textbf{Llama-3.2-1B / 3B / 8B} \\
            \midrule
            Context Length & 8,192 \\
            Batch Size & 512 \\
            Max Steps & 5,000 \\
            Warmup Steps & 0 \\
            Weight Decay & 0.1 \\
            Optimizer & AdamW \\
            \midrule
            LR Scheduler & 
            \makecell{\small Cosine Decay\\
            5e-5$\rightarrow$5e-6 / 
            2e-5$\rightarrow$2e-6 / 
            1e-5$\rightarrow$1e-6} \\
            \bottomrule
        \end{NiceTabular}
    }
    \vspace{-2.5mm}
\end{wraptable}

Based on mid-training evaluation results, the cosine decay strategy demonstrates more consistent performance. We therefore adopt it as the default scheduler for the decay stage, with hyperparameters detailed in Table~\ref{tab:octo-decay-train-config}. During the decay stage, we branch the mid-training into three distinct variants based on data composition: {\color{gblue9}{\octo-Long}} (long-reasoning data), {\color{gblue9}{\octo-Short}} (short-reasoning data), {\color{gblue9}{\octo-Hybrid}} (a mix of both) with decayed learning rate. The corresponding data mixtures are shown in Table~\ref{tab:decay-long-short-hybrid-mixture}.

\begin{table}[htbp]
  \centering
  \caption{Specific data mixture for each branch in the decay stage}
  \label{tab:decay-long-short-hybrid-mixture}
  \newcommand{\branchwidth}{0.30\textwidth}  
  \hspace*{\fill}
  \begin{minipage}[t]{0.32\textwidth}
    \centering
    \caption*{(a) Long Branch Mixture}
    \resizebox{\linewidth}{!}{
\begin{tabular}{l c}
\toprule
\textbf{Dataset} & \textbf{Weight} \\
\midrule
DCLM-Baseline & 0.05 \\
Instruction Following & 0.10 \\
MegaMath-Web-Pro & 0.55 \\
Open R1 & 0.15 \\
AM-DeepSeek-Distilled-40M & 0.15 \\
\bottomrule
\end{tabular}
}
  \end{minipage}
  \hspace*{\fill}
  \begin{minipage}[t]{\branchwidth}
    \centering
    \caption*{(b) Short Branch Mixture}
    \resizebox{\linewidth}{!}{
\begin{tabular}{l c}
\toprule
\textbf{Dataset} & \textbf{Weight} \\
\midrule
DCLM-Baseline & 0.05 \\
Instruction Following & 0.10 \\
MegaMath-Web-Pro & 0.55 \\
MegaMath-QA & 0.025 \\
OpenMathInstruct2 & 0.175 \\
NuminaMath1.5 & 0.10 \\
\bottomrule
\end{tabular}
}
  \end{minipage}
  \hspace*{\fill}
  \begin{minipage}[t]{\branchwidth}
    \centering
    \caption*{(c) Hybrid Branch Mixture}
    \resizebox{\linewidth}{!}{
\begin{tabular}{l c}
\toprule
\textbf{Dataset} & \textbf{Weight} \\
\midrule
DCLM-Baseline & 0.05 \\
Instruction Following & 0.10 \\
MegaMath-Web-Pro & 0.55 \\
OpenMathInstruct2 & 0.10 \\
NuminaMath1.5 & 0.10 \\
Open R1 & 0.10 \\
\bottomrule
\end{tabular}
}
  \end{minipage}
  \hspace*{\fill}
\end{table}

\subsection{Evaluation on \octo-Base Series}

We evaluate the performance of each branch on 13 mathematical benchmarks, alongside the original Llama base model and the model after stable-stage mid-training. As shown in Table~\ref{tab:octo_decay_1B_perf},\ref{tab:octo_decay_3B_perf},\ref{tab:octo_decay_8B_perf}, across all sizes, each OctoThinker branch demonstrates a noticeable 10\%-20\% improvement over the original base model and consistent gains over the stable-stage model. Notably, random and poor performance on challenging competition benchmarks highlights the necessity of post-training. Overall, these results reinforce our view that OctoThinker-Base series provide a strong foundation for studying RL scaling with solid reasoning capabilities.

\begin{table}[!h]
  \centering
  \caption{Evaluation results of Llama-3.2-1B and OctoThinker-1B series.}
  \label{tab:octo_decay_1B_perf}
\scalebox{0.9}
  {
    \begin{NiceTabular}{cl|c|cccc}
    \toprule
    \multicolumn{2}{c}{\multirow{2}[4]{*}{\textbf{Benchmarks}}} & \multirow{2}[4]{*}{\textbf{Llama-3.2-1B}}  & \multicolumn{4}{c}{\textbf{OctoThinker-1B-Base}}  \\
\cmidrule{4-7}    \multicolumn{2}{c}{} &   & \textbf{Stable}    & \textbf{Long} & \textbf{Hybrid} & \textbf{Short} \\
    \midrule
    {\multirow{5}{*}{Core}} 
        & {GSM8K {\tiny{(8-shot)}}} & 7.66  & 30.93 & 37.15 & 42.38 & 44.88 \\
        & {MATH500 {\tiny{(4-shot)}}} & 4.60 & 17.40  & 16.40 & 26.40 & 27.80 \\
        & {Olympiad Bench {\tiny{(4-shot)}}} & 0.89 & 2.96   & 3.41  & 5.48  & 3.85  \\
        & {AMC23 {\tiny{(0-shot)}}}  & 0.00 & 10.00 & 7.50  & 10.00 & 10.00 \\
    \cmidrule{2-7}    
        & {Average} & 3.29 & 15.32 & \textbf{16.12} & \textbf{21.07} & \textbf{21.63} \\
    \midrule
    {\multirow{10}{*}{Other}} 
        & {MATH {\tiny{(4-shot)}}} & 4.34 & 18.26 & 21.74 & 28.50 & 29.98 \\
        & {SAT MATH {\tiny{(4-shot)}}} & 12.50 & 46.88 & 31.25 & 56.25 & 46.88 \\
        & {MathQA {\tiny{(8-shot)}}} & 12.20 & 24.80 & 33.20 & 36.90 & 36.70 \\
        & {MMLU STEM {\tiny{(4-shot)}}} & 19.90 & 35.59 & 36.45 & 38.60 & 37.91 \\
        & {OCW Course {\tiny{(4-shot)}}} & 4.41  & 6.25 & 4.04  & 6.25  & 6.62  \\
        & {MAWPS {\tiny{(8-shot)}}} & 43.05 & 79.47 & 83.15 & 88.57 & 88.09 \\
        & {SVAMP {\tiny{(8-shot)}}} & 20.90 & 47.10 & 55.80 & 63.20 & 61.20 \\
        & {ASDiv {\tiny{(8-shot)}}} & 34.53 & 69.96 & 72.55 & 75.30 & 75.26 \\
        & {TabMWP {\tiny{(8-shot)}}} & 24.40 & 45.10 & 50.10 & 51.60 & 51.20 \\
\cmidrule{2-7}          
        & Average  & 19.58 & 41.49 & \textbf{43.14} & \textbf{49.46} & \textbf{48.20} \\
    \bottomrule
    \end{NiceTabular}%
}
\end{table}%

\begin{table}[!h]
  \centering
  \caption{Evaluation results of Llama-3.2-3B and OctoThinker-3B series.}
  \label{tab:octo_decay_3B_perf}
\scalebox{0.9}
  {
    \begin{NiceTabular}{cl|c|cccc}
    \toprule
    \multicolumn{2}{c}{\multirow{2}[4]{*}{\textbf{Benchmarks}}} & \multirow{2}[4]{*}{\textbf{Llama-3.2-3B}} & \multicolumn{4}{c}{\textbf{OctoThinker-3B-Base}} \\
\cmidrule{4-7}    \multicolumn{2}{c}{} &   & \textbf{Stable}  & \textbf{Long} & \textbf{Hybrid} & \textbf{Short} \\
    \midrule
    {\multirow{5}{*}{Core}} 
        & {GSM8K {\tiny{(8-shot)}}} & 30.48 & 55.95  & 56.10       & 64.37   & 63.31       \\
        & {MATH500 {\tiny{(4-shot)}}} & 7.40 & 22.40 & 25.80       & 30.80   & 31.40       \\
        & {Olympiad Bench {\tiny{(4-shot)}}} & 2.07 & 3.41 & 4.74        & 4.00    & 4.74        \\
        & {AMC23 {\tiny{(0-shot)}}}  & 2.50 & 5.00 & 7.50        & 10.00   & 2.50        \\
    \cmidrule{2-7}    
        & {Average} & 10.61 & 21.69 & \textbf{23.54}      & \textbf{27.29} & \textbf{25.49}     \\
    \midrule
    {\multirow{10}{*}{Other}} 
        & {MATH {\tiny{(4-shot)}}} & 8.24 & 24.86 & 29.98       & 31.76   & 32.70       \\
        & {SAT MATH {\tiny{(4-shot)}}} & 25.00 & 59.38 & 65.63       & 59.38   & 53.13       \\
        & {MathQA {\tiny{(8-shot)}}} & 18.20 & 39.50 & 45.40       & 47.50   & 49.80       \\
        & {MMLU STEM {\tiny{(4-shot)}}} & 38.63 & 46.32 & 48.11       & 49.73   & 48.87       \\
        & {OCW Course {\tiny{(4-shot)}}} & 5.51 & 11.40  & 11.40       & 8.46    & 9.19        \\
        & {MAWPS {\tiny{(8-shot)}}} & 79.90 & 89.69  & 91.67       & 94.24   & 93.51       \\
        & {SVAMP {\tiny{(8-shot)}}} & 52.40 & 68.40 & 69.10       & 78.00   & 77.30       \\
        & {ASDiv {\tiny{(8-shot)}}} & 60.09 & 79.59 & 79.91       & 82.80   & 82.26       \\
        & {TabMWP {\tiny{(8-shot)}}} & 48.30 & 55.60 & 56.40       & 57.80   & 60.00       \\
\cmidrule{2-7}          
        & Average  & 37.36 & 52.75 & \textbf{55.29} & \textbf{56.63}   & \textbf{56.31} \\
    \bottomrule
    \end{NiceTabular}%
}
\end{table}%

\vspace{-5pt}
\begin{table}[!h]
  \centering
  \caption{Evaluation results of Llama-3.1-8B and OctoThinker-8B series.}
  \label{tab:octo_decay_8B_perf}
\scalebox{0.9}
  {
    \begin{NiceTabular}{cl|ccccc}
    \toprule
    \multicolumn{2}{c}{\multirow{2}[4]{*}{\textbf{Benchmarks}}} & \multirow{2}[4]{*}{\textbf{Llama-3.1-8B}} & \multicolumn{4}{c}{\textbf{OctoThinker-8B-Base}} \\
\cmidrule{4-7}    \multicolumn{2}{c}{} & & \textbf{Stable}  & \textbf{Long} & \textbf{Hybrid} & \textbf{Short} \\
    \midrule
    {\multirow{5}{*}{Core}}
        & {GSM8K {\tiny{(8-shot)}}} & 55.11 & 71.27  & 72.48 & 77.41 & 77.10 \\
        & {MATH500 {\tiny{(4-shot)}}} & 20.80 & 34.40 &  37.80 & 42.60 & 38.60 \\
        & {Olympiad Bench {\tiny{(4-shot)}}} & 3.56 & 9.78  & 11.85 & 4.74  & 10.07 \\
        & {AMC23 {\tiny{(0-shot)}}}  & 5.00 & 0.00  & 5.00  & 5.00  & 7.50  \\
    \cmidrule{2-7}    
        & {Average} & 21.12 & 28.86 & \textbf{31.78} & \textbf{32.44} & \textbf{33.32} \\
    \midrule
    {\multirow{10}{*}{Other}} 
        & {MATH {\tiny{(4-shot)}}} & 21.36 & 37.00 & 41.98 & 44.82 & 38.54 \\
        & {SAT MATH {\tiny{(4-shot)}}} & 53.13 & 81.25 & 81.25 & 87.50 & 87.50 \\
        & {MathQA {\tiny{(8-shot)}}} & 36.00 & 58.20 & 62.80 & 60.50 & 62.80 \\
        & {MMLU STEM {\tiny{(4-shot)}}} & 54.44 & 62.03 & 63.75 & 64.08 & 64.38 \\
        & {OCW Course {\tiny{(4-shot)}}} & 12.87 & 18.38 & 16.18 & 15.07 & 13.97 \\
        & {MAWPS {\tiny{(8-shot)}}} & 90.75 & 93.08 &94.43 & 95.98 & 95.54 \\
        & {SVAMP {\tiny{(8-shot)}}} & 70.50 & 79.50 &82.40 & 86.10 & 86.40 \\
        & {ASDiv {\tiny{(8-shot)}}} & 72.10 & 83.79 &83.57 & 84.47 & 85.33 \\
        & {TabMWP {\tiny{(8-shot)}}} & 63.10 & 67.90  & 70.10 & 68.90 & 71.60 \\
\cmidrule{2-7}          
        & Average  & 52.69 & 64.57 & \textbf{66.27} & \textbf{67.49} & \textbf{67.34} \\
    \bottomrule
    \end{NiceTabular}%
}
\end{table}%

\clearpage

\section{OctoThinker-Zero Families: RL Scaling with Diverse Thinking Behaviors}

\begin{figure}[!h]
\centering
\vspace{-1.5pt}
\includegraphics[width=0.85\linewidth]{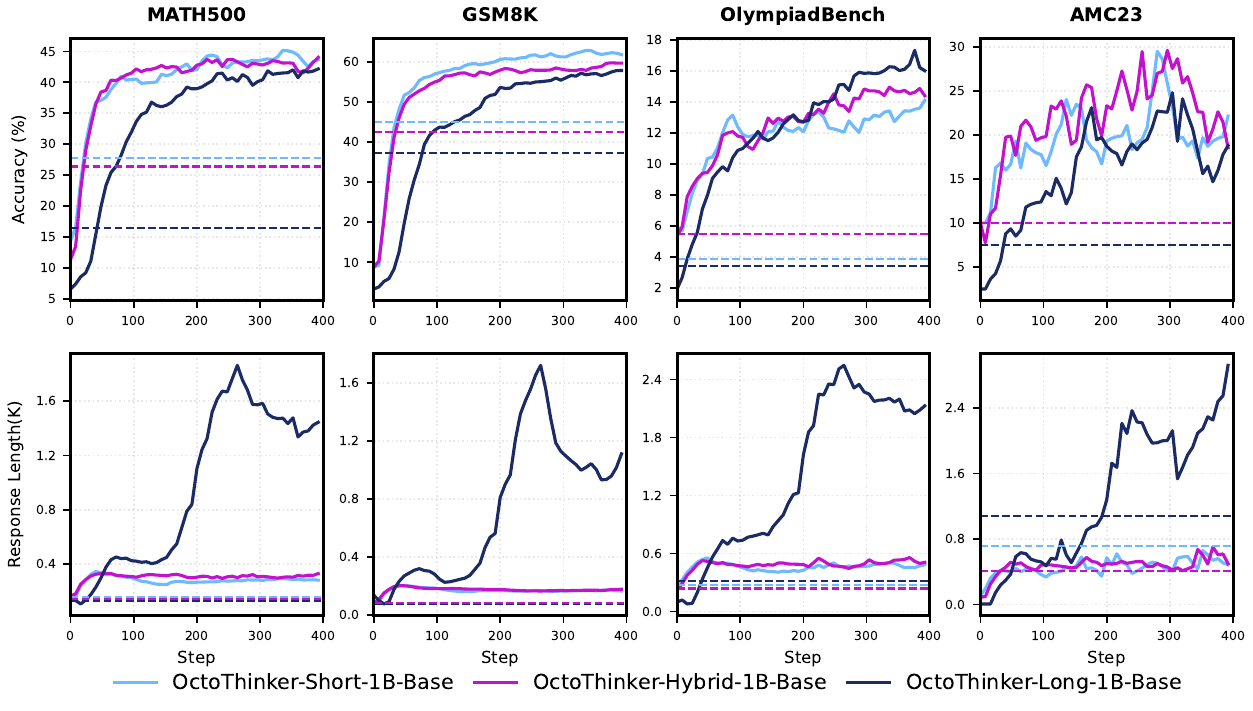}
\vspace{-1.5mm}
\caption{
The RL training dynamics across different branches for OctoThinker-1B series
}
\label{fig:octothinker_zero_1b_perf}
\end{figure}

\begin{figure}[!h]
\centering
\includegraphics[width=0.85\linewidth]{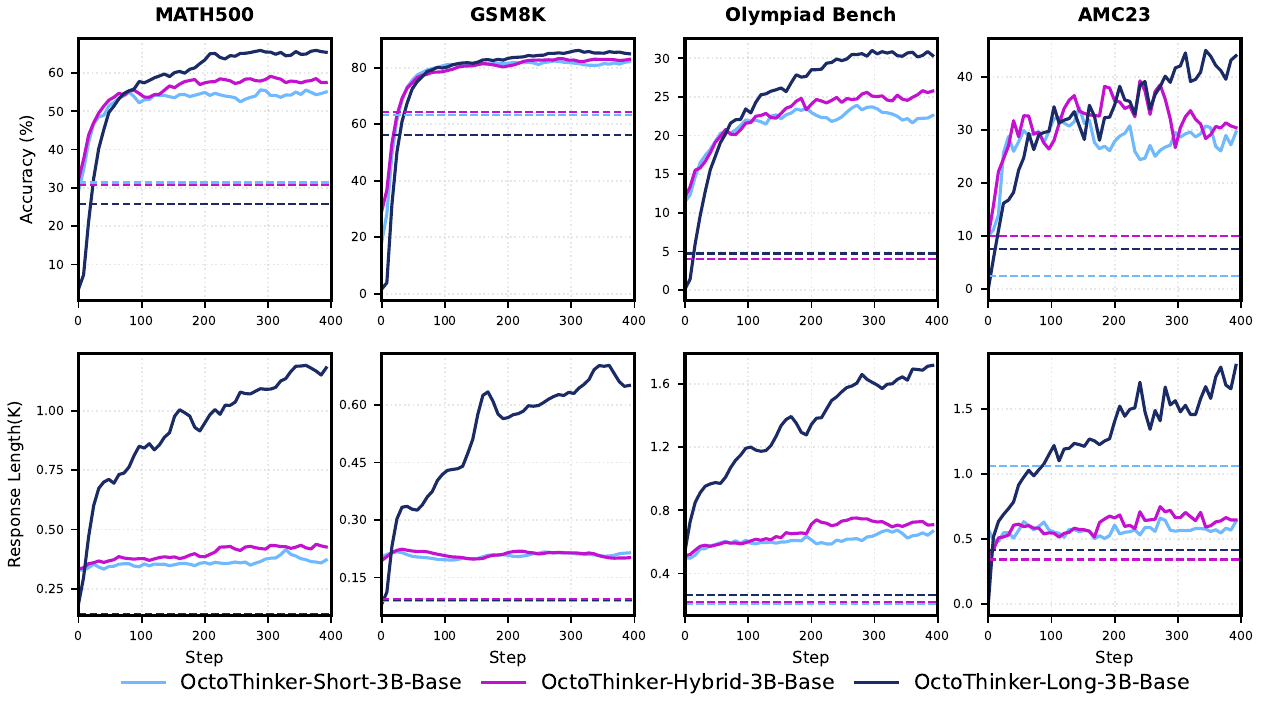}
\vspace{-1.5mm}
\caption{
The RL training dynamics across different branches for OctoThinker-3B series
}
\label{fig:octothinker_zero_3b_perf}
\end{figure}

We further train all OctoThinker base models—spanning different decay branches and model sizes (1B and 3B)—through a reinforcement learning stage, following our previously established setup. This yields a family of models optimized specifically for mathematical reasoning. As in the decay stage, the final RL-tuned models fall into three categories, each reflecting the data mixture used during decay and the distinct behaviors shaped during RL: \textHL{OctoThinker-Short-Zero}, \textHL{OctoThinker-Hybrid-Zero}, and \textHL{OctoThinker-Long-Zero}. The training dynamics of these models are shown in Figure~\ref{fig:octothinker_zero_1b_perf},\ref{fig:octothinker_zero_3b_perf}. The OctoThinker-Long branch tends to produce longer responses—within a controlled range—compared to other branches. While it slightly underperforms at the 1B scale, it demonstrates stronger performance as model size increases, such as at 3B.

\begin{figure}[!h]
\centering
\vspace{-3.5pt}
\includegraphics[width=0.95\linewidth]{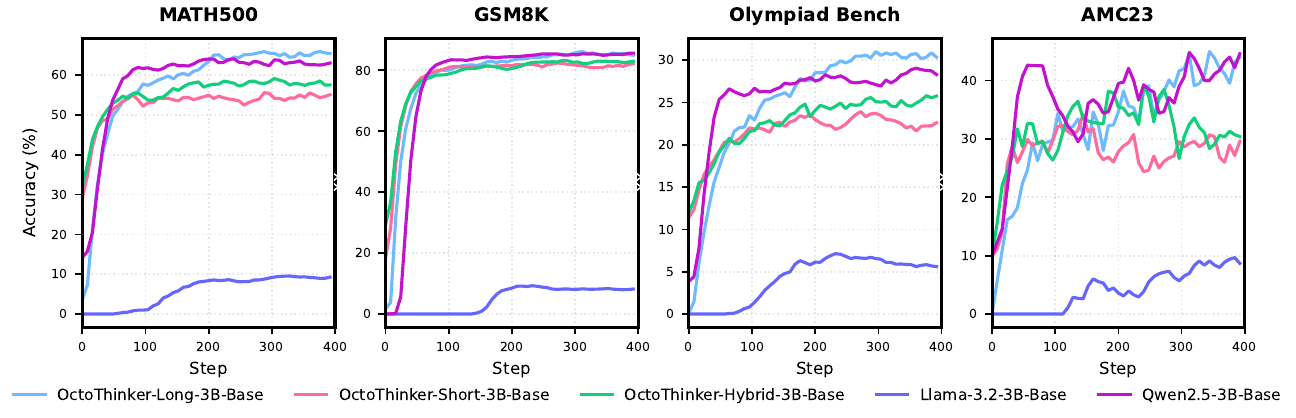}
\vspace{-1.5mm}
\caption{
RL training dynamics among Llama-3.2-3B-Base, OctoThinker series and Qwen2.5-Base.
}
\label{fig:octo_vs_llama_vs_qwen_3b}
\end{figure}

\para{OctoThinker vs. Qwen2.5} 
An important question we investigate is: \emph{To what extent can our OctoThinker models close the performance gap between the Llama-3.2 series and the stronger Qwen2.5 models in the RL setting?} To address this, we compare three 3B-scale base models: Llama-3.2-3B-Base, OctoThinker-Long-3B-Base, and Qwen2.5-3B-Base. As illustrated in Figure~\ref{fig:octo_vs_llama_vs_qwen_3b}, during the reinforcement learning phase, OctoThinker-Long-3B consistently outperforms the original Llama-3.2-3B model. Remarkably, it reaches performance on par with Qwen2.5-3B, a model known for its strong reasoning capabilities and extensive pre-training, while the hybrid and short branches are marginally inferior, especially on challenging benchmarks. Overall, these results highlight the substantial gains introduced by our mid-training strategy and confirm that OctoThinker effectively narrows the performance gap, elevating Llama-3.2 models to a new level of competitiveness in mathematical reasoning tasks.

\section{Related Works}

\para{Understanding RL along with Language Models} Large-scale RL has driven the major advances in language models on reasoning-intensive tasks, such as competition-level math (e.g., AIME),  exemplified by OpenAI's o1~\cite{Contributors2024OpenAIo1}, o3~\cite{openai2025o3} and DeepSeek's R1~\cite{guo2025deepseek}. A wave of follow-up studies~\citeg{zeng2025simplerl,hu2025open,DBLP:journals/corr/abs-2503-14476-dapo,deepscaler2025} explored RL on smaller language models (less than 100B parameters), yet these successes are overwhelmingly limited to Qwen family. In contrast, replicating such results on other major model families—e.g., Llama-has proven difficult~\cite{gandhi2025cognitive,liu2025understanding}. The opacity of pre-training pipelines hinders our understanding of how pre-training interacts with RL scaling, prompting some unconventional investigations~\cite{DBLP:journals/corr/abs-2504-20571-oneshot-rlvr,Shao2025SpuriousRR}. For instance, \citet{DBLP:journals/corr/abs-2504-20571-oneshot-rlvr} showed that even one-shot prompting can enhance reasoning in Qwen, but yields minimal gains in Llama. The underlying science remains essential but largely unexplored. Our work takes a step toward filling this gap by performing controlled mid-training interventions on Llama models, revealing key factors that enable effective RL scaling. Building on these insights, we introduce OctoThinker via a two-stage mid-training strategy (over 200B tokens), followed by RL training, yielding models that match Qwen’s performance at the same scale.

\para{Curation of Math Pre-training Corpora} Pre-training corpora are foundational to language models, especially for math reasoning tasks where large-scale mid-training is infeasible without high-quality, domain-specific data. Early open-source efforts—such as OpenWebMath~\cite{paster2024openwebmath}, MathPile~\cite{wang2024mathpile}, InfiMM-Web-Math~\cite{han2024infimmwebmathb}, and FineMath~\cite{DBLP:journals/corr/abs-2502-02737-smollm2}—have made meaningful progress but remain constrained in scale, typically under 100B tokens. The release of MegaMath~\cite{Zhou2025MegaMath} marked a turning point, enabling scalable mid-training in this work. Building on its foundation, we curated a new reasoning-intensive and carefully refined math corpus, \mmwebpromax, which exceeds 70B tokens and matches the quality of MegaMath-Web-Pro. This corpus powers the first stage of our mid-training of OctoThinker and will be released to support the broader open-source community.

\section{Conclusion}

In this work, we investigate why base models like Llama and Qwen exhibit divergent behaviors during reinforcement learning for reasoning and demonstrated that mid-training can play a decisive role. Our findings show that high-quality, reasoning-intensive corpora—especially those like MegaMath-Web-Pro—can substantially improve RL stability and effectiveness. Building on these insights, we introduce a two-stage mid-training strategy that transforms Llama into a more RL-scalable foundation model. The resulting OctoThinker models achieve strong performance across mathematical reasoning tasks, closing the gap with RL-friendly model families. We hope this work provides a foundation for designing future base models better aligned with the demands of reasoning-centric RL.

\section*{Future Work}

We will actively explore more in the future, include: (1) curating higher-quality math corpora to further enhance mid-training; (2) designing RL-friendly base models using open recipes without distillation from those powerful long CoT reasoning models; (3) disentangling QA format and content to better understand their individual contributions; and (4) extending the OctoThinker families with additional branches, such as \emph{tool-integrated reasoning}. We believe these efforts will provide deeper insights into the interplay between pre-training and reinforcement learning.

\bibliography{main}

\clearpage
\newpage
\appendix
\section*{\centering \LARGE{Appendix}}
\vspace{-2.5mm}
\begin{figure}[htbp]
\begin{tcolorbox}[
colback=gblue9!5,
colframe=gblue9!75,
left=2mm, right=2mm,title=\textcolor{white}{\textbf{Scoring Prompt of Usefulness  for Studying Mathematics}}]
Evaluate the following text extract for its potential usefulness for studying mathematics up to high school and early undergraduate levels. Use the following 5-point scoring system described below. Points are accumulated based on the satisfaction of each criterion:

- Add 1 point if the extract contains some mathematical content, even if it's not very useful for studying, or if it contains non-academic content such as advertisements and generated pages for converting weight and currencies.

- Add another point if the extract touches on mathematical topics, even if it's poorly written if it's too complex such as an academic paper that is too advanced. 

- Award a third point if the extract demonstrates problem solving or logical reasoning in a mathematical context, even if it lacks step-by-step explanations.

- Grant a fourth point if the extract is at an appropriate level (up to high school and early undergraduate levels) and contains clear mathematical deductions and step-by-step solutions to mathematical problems. It should be similar to a chapter from a textbook or a tutorial.

- Give a fifth point if the extract is outstanding in its educational value for teaching and studying mathematics in middle school and high school. It should include very detailed and easy to follow explanations.

Question-answer formats (e.g., from educational websites or forums) are acceptable if they meet the criteria.

The text extract:
``
<document>
''

After examining the extract:

- Briefly justify your total score, up to 100 words.

- Conclude with the score using the format: Final score: <total points>.

\end{tcolorbox}
\vspace{-2.5mm}
\caption{Scoring prompt in FineMath~\citep{DBLP:journals/corr/abs-2502-02737-smollm2} of usefulness for studying mathematics.}
\label{app-fig:finemath-prompt}
\end{figure}
\vspace{1.0mm}
\begin{figure}[H]
\begin{tcolorbox}[
colback=gblue9!5,
colframe=gblue9!75,
left=2mm, right=2mm,title=\textcolor{white}{\textbf{Web Text Refinement Prompt}}]
Task: 

- Carefully analyze the provided text to extract key facts, concrete details, important numbers, and core concepts. 

- Remove any irrelevant or noisy information, and reorganize the content into a logically structured, information-dense, and concise version that is easy to learn from. Output only the refined text.

- Strive to maintain the original length as much as possible (avoid excessive shortening).

- Refine multiple choice questions and answers if any.

Text:

<EXAMPLE>

Just output the refined text, no other text.

\end{tcolorbox}
\vspace{-2.5mm}
\caption{Web text refinement prompt used in MegaMath-Web-Pro~\citep{Zhou2025MegaMath}}
\label{app-fig:text-refining-prompt}
\end{figure}

\begin{figure}[!htbp]
\centering
\includegraphics[width=0.8\textwidth]{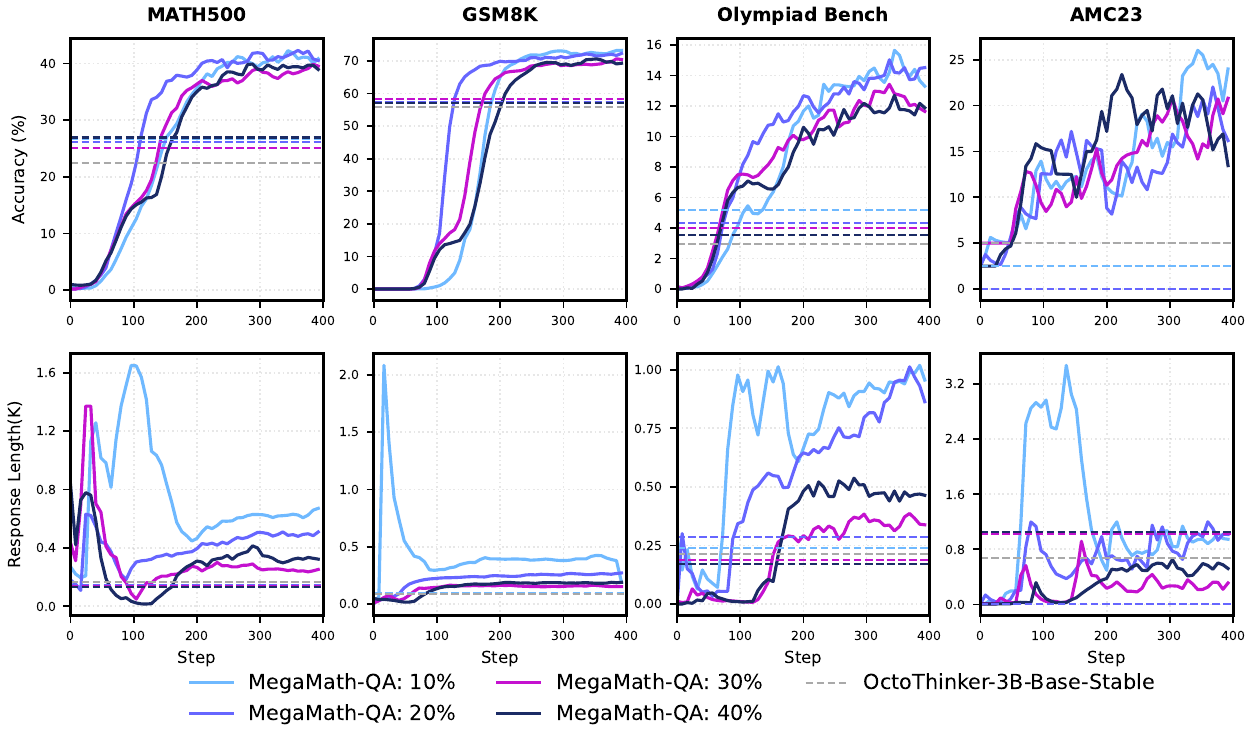}

\includegraphics[width=0.8\textwidth]{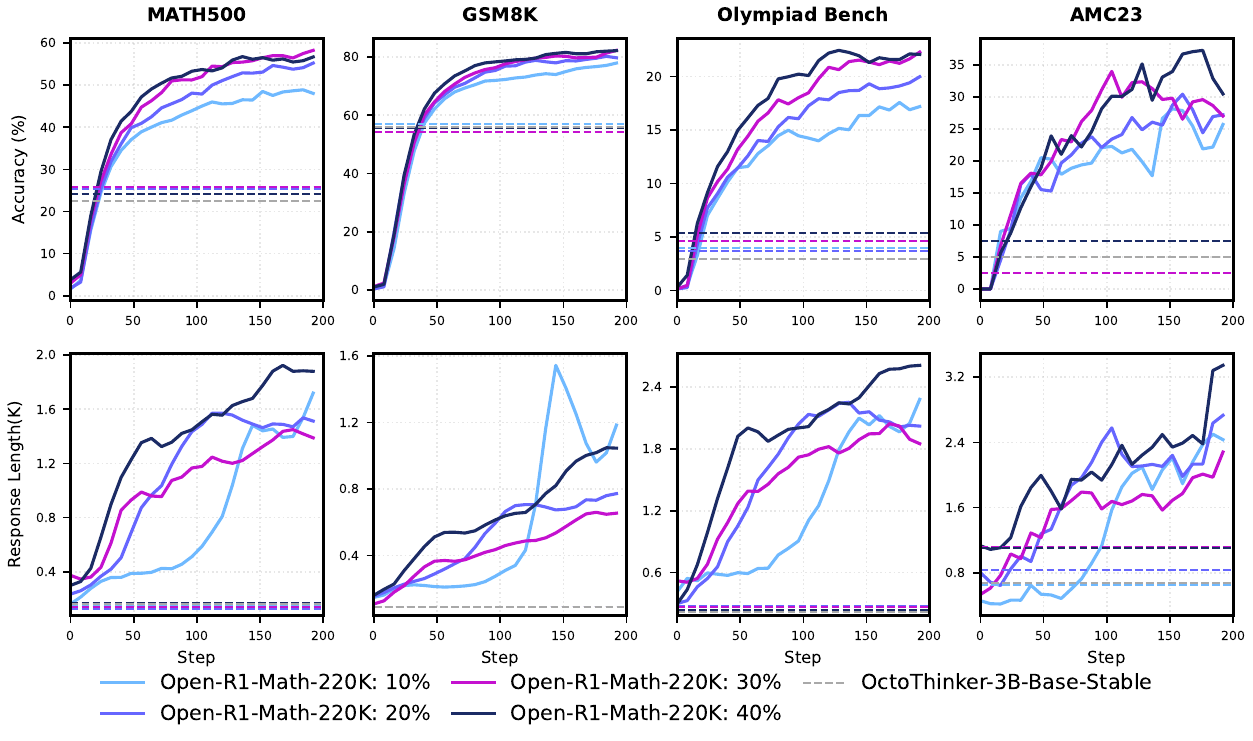}

\includegraphics[width=0.8\textwidth]{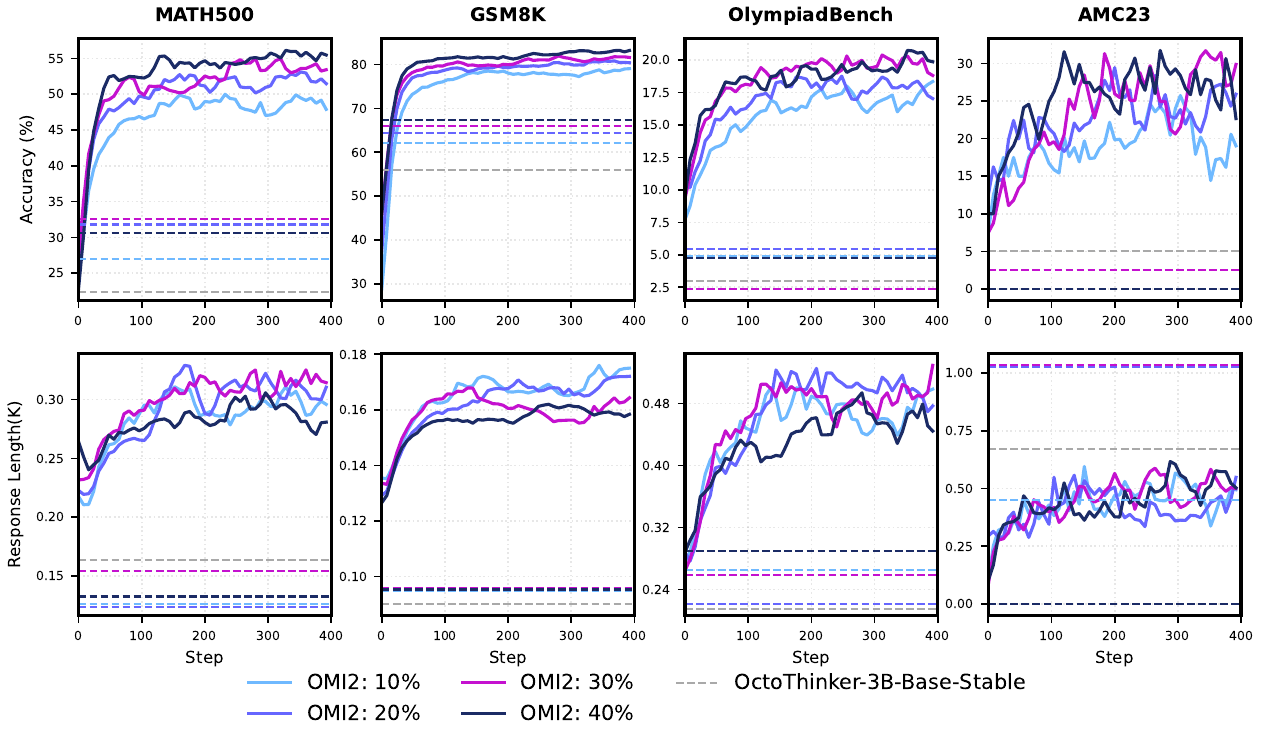}
\vspace{-3mm}
\caption{RL dynamics under different QA datasets and mixing ratios during the decay stage.}
\label{fig:decay_ablation_qa_ratio}
\end{figure}

\end{document}